\documentclass{article}

\usepackage{microtype}
\usepackage{graphicx}
\usepackage{subfigure}

\usepackage[utf8]{inputenc}
\usepackage{booktabs}       
\usepackage{multirow}       
\usepackage{graphicx}       
\usepackage[table]{xcolor}  
\usepackage{arydshln}       
\usepackage{makecell}       

\usepackage[accepted]{icml2026} 

\usepackage{hyperref}
\usepackage{xcolor}
\usepackage{enumitem}
\usepackage{pifont}

\definecolor{mydarkblue}{rgb}{0,0.08,0.45}

\hypersetup{
  colorlinks=true,
  linkcolor=mydarkblue,
  citecolor=mydarkblue,
  filecolor=mydarkblue,
  urlcolor=mydarkblue,
}

\usepackage{amsmath}
\usepackage{amssymb}
\usepackage{mathtools}
\usepackage{amsthm}
\usepackage{algorithm}
\usepackage{algorithmic}
\usepackage{xspace}    
\usepackage{textcomp} 

\usepackage[capitalize,noabbrev]{cleveref}
\crefname{equation}{\textcolor{red}{equation}}{\textcolor{red}{equations}}
\Crefname{equation}{\textcolor{red}{Equation}}{\textcolor{red}{Equations}}
\usepackage[most]{tcolorbox} 

\newcommand{\infobox}[1]{
    \vspace{-0.13cm}
    \begin{tcolorbox}[
        colback=white!90!gray,     
        colframe=teal!50!black,    
        arc=5pt,                   
        boxsep=5pt,                
        left=5pt, right=10pt,      
        top=2pt, bottom=3pt,       
        boxrule=0.8pt,             
        fontupper=\itshape         
    ]
    \vspace{-0.1cm}
         \textit{#1}
    \vspace{-0.2cm}
    \end{tcolorbox}
    \vspace{-0.15cm}
}

\newcommand{\ourmethod}{{\fontfamily{lmtt}\selectfont \textbf{AR-MAP}}\xspace}

\theoremstyle{plain}
\newtheorem{theorem}{Theorem}[section]
\newtheorem{proposition}[theorem]{Proposition}

\theoremstyle{definition}

\theoremstyle{remark}

\definecolor{highlightblue}{rgb}{0.88, 0.95, 1.0}

\usepackage{colortbl} 
\definecolor{headergray}{gray}{0.95} 
\definecolor{highlight}{HTML}{EAF2FF} 

\usepackage[textsize=tiny]{todonotes}

\icmltitlerunning{AR-MAP: Are Autoregressive Large Language Models Implicit Teachers for Diffusion Large Language Models?}

\begin{document}

\twocolumn[
\icmltitle{AR-MAP: Are Autoregressive Large Language Models Implicit Teachers for Diffusion Large Language Models? }

\icmlsetsymbol{equal}{*}
\icmlsetsymbol{leader}{\textdagger}      
\icmlsetsymbol{comp}{\textdaggerdbl}     

\begin{icmlauthorlist}
  \icmlauthor{Liang Lin}{yyy,comp}  
  \icmlauthor{Feng Xiong}{yyy}
  \icmlauthor{Zengbin Wang}{yyy}
  \icmlauthor{Kun Wang}{sch}
  \icmlauthor{Junhao Dong}{sch}
  \icmlauthor{Xuecai Hu}{yyy,leader}      
  \icmlauthor{Yong Wang}{yyy,leader}
  \icmlauthor{Xiangxiang Chu}{yyy}
\end{icmlauthorlist}

\icmlaffiliation{yyy}{AMAP, Alibaba Group.}
\icmlaffiliation{sch}{Nanyang Technological University} 

\icmlkeywords{Machine Learning, ICML}

\vskip 0.3in
]


\printAffiliationsAndNotice{%
  \textsuperscript{\textdagger}Project leads. \hspace{1em}
  \textsuperscript{\textdaggerdbl}Work done during the internship at AMAP, Alibaba Group.
}

\begin{abstract}

Diffusion Large Language Models (DLLMs) have emerged as a powerful alternative to autoregressive models, enabling parallel token generation across multiple positions. However, preference alignment of DLLMs remains challenging due to high variance introduced by Evidence Lower Bound (ELBO)-based likelihood estimation. In this work, we propose AR-MAP, a novel transfer learning framework that leverages preference-aligned autoregressive LLMs (AR-LLMs) as implicit teachers for DLLM alignment. We reveal that DLLMs can effectively absorb alignment knowledge from AR-LLMs through simple weight scaling, exploiting the shared architectural structure between these divergent generation paradigms.  Crucially, our approach circumvents the high variance and computational overhead of direct DLLM alignment and comprehensive experiments across diverse preference alignment tasks demonstrate that AR-MAP achieves competitive or superior performance compared to existing DLLM-specific alignment methods, achieving 69.08\% average score across all tasks and models.  Our  Code is available at \url{https://github.com/AMAP-ML/AR-MAP}.

\end{abstract}
\section{Introduction}
\begin{figure}
    \centering
    \includegraphics[width=1\linewidth]{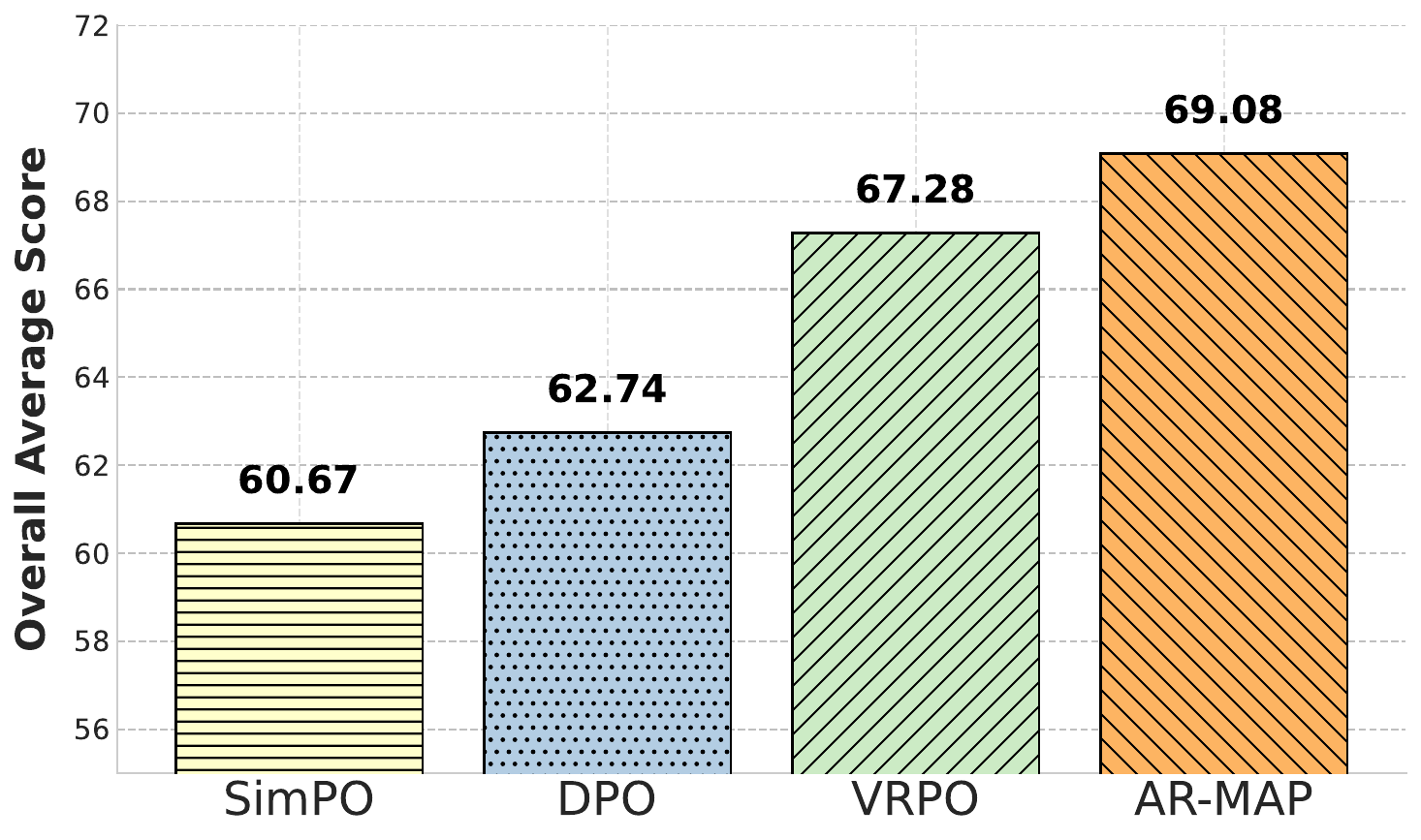}
    \caption{The average performance of \ourmethod and baseline method on all tasks and models.}
    \label{fig:intro}
    \vspace{-1.5em}
\end{figure}


The advancement of Diffusion Large Language Models (DLLMs)~\citep{ye2023diffusion,nie2024scaling,zhang2025survey,yu2025dapo} has disrupted the autoregressive (AR)~\citep{grattafiori2024llama,ahmed2025qwen,wang2025comprehensive,li2025adacurl} paradigm of left-to-right generation, making it possible to generate multiple tokens at different positions simultaneously. This architectural advancement enables models to incorporate bidirectional contextual information during inference, thereby facilitating the parallel generation of multiple tokens and yielding a substantial improvement in generation efficiency. 
Notably, recent open-source models such as \texttt{LLaDA}~\citep{zhu2025llada,bie2025llada2}, \texttt{Dream}~\citep{ye2025dream,xie2025dream}, and \texttt{SDAR}~\citep{cheng2025sdar} have demonstrated performance that is competitive with prevailing autoregressive models.

Building upon this, contemporary research has concentrated on enhancing advanced reasoning capabilities in complex tasks \citep{wang2025d2,wang2025revolutionizing,tang2025wd1,huang2025reinforcing,ji2025tree,chu2025gpg}, accelerating inference processes via sparsification or parallelization \citep{huang2025pc,song2025sparse,wu2025fast}, and optimizing instruction alignment through methods such as human feedback \citep{ziegler2019fine,lee2025instructpatentgpt,xiong2025hs,ji2026thinking}. Despite these significant advances, the generative paradigm grounded in diffusion or flow models remains hampered by two persistent challenges inherent to Evidence Lower Bound (ELBO) likelihood estimation: \textbf{uncertainties} and \textbf{high variance}. To mitigate uncertainty, diffu-GRPO \citep{xie2025step} approximates the ELBO via a single forward pass for fully masked completion, while d2-stepMerge \citep{wang2025d2} estimates likelihood by accumulating conditional transition probabilities across multiple denoising steps; however, both approaches rely on rigid, pre-defined assumptions regarding sequence order to approximate the complete diffusion trajectory, often failing to capture the true data distribution and leading to estimation bias. Furthermore, to address high variance, VRPO \citep{zhu2025llada} increases the decoding time steps for masked token sampling to stabilize outputs, yet this inevitably incurs prohibitive computational overhead by sacrificing inference efficiency. Consequently, achieving an optimal equilibrium between generation quality and computational cost remains an unresolved core challenge in this domain.

Though these methods have brought significant insights, they all operate within the inherent architectural constraints of DLLMs. As the latest generation of large diffusion models increasingly adopts the paradigm of transitioning from AR-to-diffusion (e.g., \textit{Dream}, \textit{SDAR}, \textit{LLaDA 2.0}), there is a growing consensus that AR-LLMs \textbf{inherently retain strong capabilities and knowledge.} This motivates us to explore the following research question(RQ) :
  \infobox{
RQ1: Can knowledge from preference-aligned AR-LLMs, which are trained efficiently and deterministically, transfer to DLLMs?
  }

To investigate whether there is a weight correlation between AR-LLMs and DLLMs, we propose \ourmethod. Inspired by task vectors~\citep{ilharco2022editing,sun2025task,akiba2025evolutionary,ma2025and}, that is, the fine-tuning weight difference after fine-tuning in a specific domain can represent the inspiration of specific task features, we investigate for the first time whether the weight differences derived from AR-LLMs to DLLMs can be treated as task vectors. To be more specific, we propose a new paradigm for efficient and stable alignment training, together with a simple but effective weight transfer algorithm: by performing autoregressive DPO on AR-LLMs to exploit their training efficiency, we compute task vectors from the fine-tuned weight differences and scale them according to reward modeling fit. As shown in Figure~\ref{fig:intro}, our experiments across 6 tasks demonstrate that \ourmethod has excellent generalization ability and increases the performance even better than the baseline \texttt{VRPO}, which is trained directly on DLLMs to minimize variance as much as possible, thereby proving its effectiveness. Furthermore, we also find that the degree of absorption enhancement preferred by DLLMs is related to the training effectiveness of AR-LLMs.
Our key contributions are summarized as the following three aspects:
\begin{itemize}
[leftmargin=*,itemsep=3pt,topsep=0.pt,parsep=0pt]
    \item \textbf{Revealed Weight Mapping:} We have revealed for the first time the existence of weight mapping between AR-LLMs and DLLMs. 

    \item \textbf{Efficient Alignment:} We propose \ourmethod, an alignment framework through training AR-LLMs and simple but effective weight transfer from AR-LLMs to DLLMs.
 
    \item \textbf{Exploratory Findings In Experiments:} Through extensive experiments, we have revealed the preference conversion pattern of AR to DLLM.
\end{itemize}

\section{Preliminaries} \label{section:prel}

\textbf{Diffusion Large Language Models.} 
The inference process of LLMs is characterized by autoregressive generation~\citep{xiong2024autoregressive,deng2025following}, where each token is predicted conditionally upon the sequence of preceding tokens, thus constructing the output in a sequential, left-to-right manner. 
In stark contrast, DLLMs~\citep{bie2025llada2,nie2025large,ye2025dream,cheng2025sdar} operate on a different principle. 
They employ a forward-reverse framework built upon a dual-process mechanism: a fixed forward process systematically corrupts a clean input from $t=0$ into a fully masked state at $t=1$. 
A single generation step involves predicting the original tokens based on a globally corrupted version of the sequence, which can be formally represented as:
\begin{equation} \label{eq:2}
\mathbf{p}(\mathbf{y}|\mathbf{x}) = \mathbb{E}_{\substack{t \sim \mathcal{U}[0,1] \\ \mathbf{y}_t \sim \mathbf{q}(\cdot|\mathbf{y},\mathbf{x})}} \left[ \prod_{j: y_{t,j}=\texttt{[MASK]}} \mathbf{p}_\theta(y_j | \mathbf{y}_t, t, \mathbf{x}) \right].
\end{equation}
In this formulation, the overall sequence probability $\mathbf{p}_\theta(\mathbf{y}|\mathbf{x})$ is an expectation $\mathbb{E}$ over uniformly sampled time steps $t$ and corresponding noisy sequences $\mathbf{y}_t$ generated by the forward process $\mathbf{q}$. The core of the generation lies in the product term, which is computed over all positions $j$ where the token has been corrupted into a special \texttt{[MASK]} token which serves as the concrete realization of noise injection.


\textbf{DPO in DLLMs.} DPO~\citep{rafailov2023direct} directly fine-tunes a policy model, $\pi_\theta$, on a human preference dataset of $(x, y_w, y_l)$. 
As shown in Equation~\ref{eq3}, the loss function maximizes the likelihood of the policy model $\pi_\theta$ preferring $y_w$ over $y_l$. It works by \textbf{increasing the log-probability ratio} for the winning response relative to the initial reference model, $\pi_{\text{0}}$, while decreasing this ratio for the losing response. This directly optimizes $\pi_\theta$ without an explicit reward model, where the  $\beta$ scales the preference strength. 
\begin{equation} \label{eq3}
\mathcal{L}_{\pi_\theta} = - \left[ \log \sigma\left(\beta \log \frac{\pi_\theta(y_w|x)}{\pi_{\text{0}}(y_w|x)} - \beta \log \frac{\pi_\theta(y_l|x)}{\pi_{\text{0}}(y_l|x)}\right) \right].
\end{equation} 

However, in DLLMs, the exact log-likelihood $\log p(\mathbf{y}|\mathbf{x})$ is intractable due to the integral over all corruption paths in Equation~\ref{eq:2}. Instead, the \textbf{Evidence Lower Bound (ELBO)} is used as a surrogate:  
\begin{equation} \label{eq4}  
B_\pi(y|x) \triangleq \mathbb{E}_{t,\mathbf{y}_t}\left[\ell_\pi(\mathbf{y}_t,t,y|x)\right]\leq\log\pi(y|x),  
\end{equation}  
where $t \in [0,1]$ is a diffusion time step, $\mathbf{y}_t$ is a masked sequence generated by the forward process $q(\cdot|y,x)$, and $\ell_\pi$ is the per-step mask-prediction loss. The ELBO introduces a nested expectation over time steps and masked data, requiring doubly stochastic estimation. To adapt DPO to DLLMs, the log-probability ratios in Equation~\ref{eq3} are replaced with ELBO differences:  
\begin{equation} \label{eq5}
\mathcal{L}_{\pi_\theta}^{\text{ELBO}} = - \log \sigma\left( \beta \log \textcolor{red}{\frac{B_{\pi_\theta}(y_w|x)}{B_{\pi_0}(y_w|x)}} - \beta \log \textcolor{red}{\frac{B_{\pi_\theta}(y_l|x)}{B_{\pi_0}(y_l|x)}} \right).
\end{equation}
This substitution inherits the high variance of ELBO estimators, which arises from the doubly stochastic Monte Carlo approximation of the nested expectations.
\section{Methodology}

In Section \ref{section:prel},  we have discussed the uncertainties and high variance introduced by ELBO-based likelihood estimation in DLLMs.  In this section, we explore whether the stability and efficiency of AR-LLMs training be used to help align DLLMs' preferences .We introduce \ourmethod, a systematic framework to investigate the mapping relationship between DLLMs and homologous AR.

\subsection{RQ1: Can Weight Transfer from AR-LLMs to DLLMs?}
To our knowledge, we present the \textbf{first} study on transferring weights between divergent reasoning architectures. We will conduct feasibility analysis and verification from the following two aspects :
\begin{table}[htbp]
    \centering
    \renewcommand{\arraystretch}{1.0}
    \setlength{\tabcolsep}{3pt}
    \setlength\dashlinedash{2pt}
    \setlength\dashlinegap{2pt}

    \caption{Examples of parameter comparison between AR-LLMs and their DLLMs obtained via continuous pre-training (CPT). The blue rows indicate the corresponding DLLMs, while the white rows represent AR-LLMs.}
    \label{tab:ar_dllm_params}

    \begin{tabular}{lccc}
        \toprule
        \textbf{Model} & \textbf{head\_num} & \textbf{hidden\_size} & \textbf{layers} \\
        \midrule

        Qwen3-4B-Base & 32 & 2560 & 36 \\
        \rowcolor{highlightblue} SDAR-4B-Instruct & 32 & 2560 & 36 \\
        \hdashline
        Qwen3-8B-Base & 32 & 4096 & 36 \\
        \rowcolor{highlightblue} SDAR-8B-Instruct & 32 & 4096 & 36 \\
        \hdashline
        Qwen2.5-7B & 28 & 3584 & 28 \\
        \rowcolor{highlightblue} Dream-7B-Instruct & 28 & 3584 & 28 \\
        \hdashline
        Ling-mini-2.0-base & 16 & 2048 & 20 \\
        \rowcolor{highlightblue} LLaDA2.0-mini & 16 & 2048 & 20 \\

        \bottomrule
    \end{tabular}
\end{table}

\textbf{Model Architecture.} In previous studies, Model Merging~\citep{yadav2023ties,yu2024language,hu2025can} has demonstrated that AR-LLMs can be combined through weight interpolation. Given a base model $\mathbf{W}_0$, and two fine-tuned models with the same architecture $\mathbf{W}_1$ (task-1 specialized) and $\mathbf{W}_2$ (task-2 specialized), the merged model is defined as:
\begin{equation}
\mathbf{W}_{\text{new}} = \mathbf{W}_0 + \alpha (\mathbf{W}_1 - \mathbf{W}_0) + \beta (\mathbf{W}_2 - \mathbf{W}_0), \quad 
\label{eq:merge}
\end{equation}
Here, $\alpha$ and $\beta$ control the contribution of each task-specific weight delta to the final model. This framework enables fine-grained control over task performance trade-offs. Therefore, as long as there is the same model skeleton, merging may take effect. As shown in Table \ref{tab:ar_dllm_params}, we can observe that the transformation from AR-LLMs~\citep{team2024qwen2,yang2025qwen3,team2025every} to DLLMs has not made any modifications to the key architecture required for merging.


\textbf{Theoretical Analysis.} Following \texttt{Task Arithmetic}~\citep{ilharco2022editing}, we first define the alignment task vector $\tau_{\mathrm{pref}}$ and the diffusion adaptation vector $\tau_{\mathrm{diffusion}}$ relative to the base autoregressive weights $\mathbf{W}_{\mathrm{AR}}$:
\begin{equation}
    \tau_{\mathrm{pref}} = \mathbf{W}_{\mathrm{AR}}^{\mathrm{aligned}} - \mathbf{W}_{\mathrm{AR}}, \quad
    \tau_{\mathrm{diffusion}} = \mathbf{W}_{\mathrm{DLLM}} - \mathbf{W}_{\mathrm{AR}},
    \label{eq:6}
\end{equation}
where $\mathbf{W}_{\mathrm{AR}}^{\mathrm{aligned}}$ and $\mathbf{W}_{\mathrm{DLLM}}$ denote the weights of the aligned AR-LLMs and the DLLMs, respectively. Here, $\tau_{\mathrm{diffusion}}$ can be interpreted as a structural task vector that enables the autoregressive model to comprehend the denoising-based inference paradigm. This framework suggests that the alignment knowledge and diffusion reasoning are additive in the shared parameter space, allowing DLLMs to effectively absorb the alignment state via $\mathbf{W}_{\mathrm{DLLM}}^{\mathrm{aligned}} \approx \mathbf{W}_{\mathrm{AR}} + \tau_{\mathrm{diffusion}} + \gamma \cdot \tau_{\mathrm{pref}}$.


\begin{figure}
    \centering
    \includegraphics[width=1.0\linewidth]{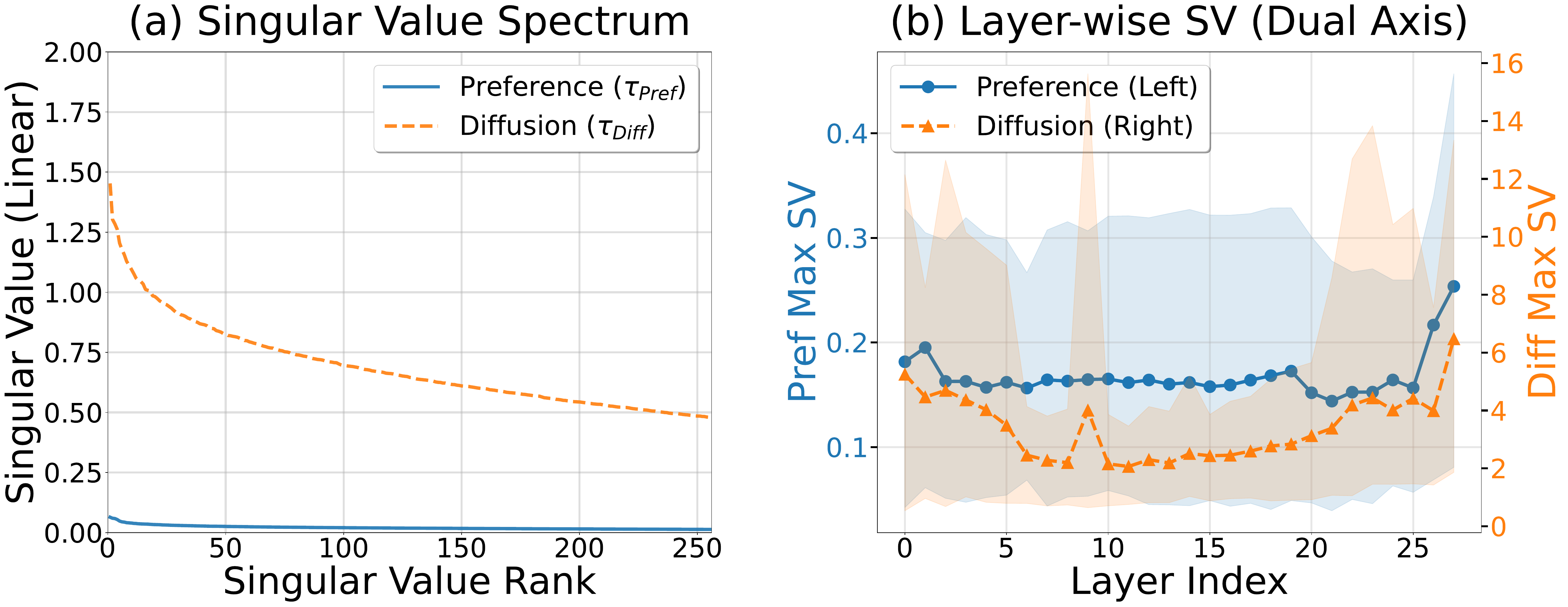}
    \caption{Singular value analysis of task vectors. \textbf{(a)} Singular value spectrum in one MLP layer. \textbf{(b)} Layer-wise maximum singular values where solid/dashed lines denote means and \textbf{shaded areas represent the min-max range}.}
    \label{fig:layer_svd}
\end{figure}

\subsection{RQ2: Discovery of Weight Scaling Law} \label{section:RQ2}
In this section, we delve into the alignment issue in model merging by identifying parameters associated with $\tau_{\mathrm{diffusion}}$ and $\tau_{\mathrm{pref}}$. We try to answer the following RQ:

\infobox{
\textbf{RQ2:} Can weights be directly transferred and do they need to be scaled accordingly? 
}

To answer this, we analyze the magnitude of parameter updates via \textbf{Singular Value Decomposition (SVD)}~\citep{stewart1993early}. By treating each task vector $\tau \in \mathbb{R}^{d_{\text{in}} \times d_{\text{out}}}$ as a linear operator, we decompose it into its spectral components:
\begin{equation}
\label{eq:svd}
\tau = U \Sigma V^\top = \sum_{i=1}^{r} \sigma_i \mathbf{u}_i \mathbf{v}_i^\top, \quad \text{s.t.} \quad \sigma_1 \ge \dots \ge \sigma_r > 0,
\end{equation}
where $\mathbf{u}_i$ and $\mathbf{v}_i$ are the left and right singular vectors representing the \textbf{principal directions} of the weight update, and $\sigma_i$ represents the \textbf{spectral energy} (scaling magnitude) along these directions. While $\mathbf{u}_i$ and $\mathbf{v}_i$ encode the semantic orientation of the task, the \textbf{spectral norm} $\|\tau\|_2 \triangleq \sigma_1$ dictates the maximal impact of the interference.

We use the \texttt{Llama-3.1-Nemotron-70B-Reward-HF} reward model\footnote{\url{https://huggingface.co/nvidia/Llama-3.1-Nemotron-70B-Reward-HF}} on \texttt{HelpSteer2}~\citep{wang2024helpsteer} to score 8 responses sampled from \texttt{Qwen2.5-7B}~\citep{team2024qwen2} for each task. 
The highest and lowest-scoring outputs are used to construct preference pairs, based on which we define the task vector $\tau_{\mathrm{pref}}$ as the weight update resulting from DPO training, and for $\tau_{\mathrm{diff}}$, we use \texttt{Dream-7B-Instruct} as backbone.
 As illustrated in Figure \ref{fig:layer_svd}, explicitly measuring these properties reveals a distinct "spectral gap": the spectral norms of diffusion updates ($\tau_{\mathrm{diffusion}}$) are orders of magnitude larger than those of preference alignment ($\tau_{\mathrm{pref}}$). This magnitude disparity leads to a phenomenon we term "Spectral Shadowing."  The detail is formalized below.

\begin{proposition}[Spectral Shadowing in Heterogeneous Task Merging]
Let $\tau_{\mathrm{diffusion}}, \tau_{\mathrm{pref}}$ be the task vectors defined in Eq. (\ref{eq:6}). Given the dominance of the diffusion adaptation magnitude, we assume the condition $\|\tau_{\mathrm{pref}}\|_2 \le \epsilon \|\tau_{\mathrm{diffusion}}\|_2$ holds for a scalar $\epsilon \ll 1$.

Under weighted summation (with scaling factor $\gamma$), the relative perturbation to the diffusion manifold is strictly controlled. By applying Weyl's inequality to the combined spectrum, we derive the upper bound for the perturbed weight space:
\vspace{-0.8em}
\begin{equation}
\begin{aligned}
\|\tau_{\mathrm{diffusion}} + \gamma \tau_{\mathrm{pref}}\|_2 
&\le \|\tau_{\mathrm{diffusion}}\|_2 + \|\gamma \tau_{\mathrm{pref}}\|_2 \\
&\le \|\tau_{\mathrm{diffusion}}\|_2 + \gamma \|\tau_{\mathrm{pref}}\|_2 \\
&\le (1+\gamma\epsilon)\|\tau_{\mathrm{diffusion}}\|_2.
\end{aligned}
\end{equation}
\label{proposition:1}
\vspace{-0.8em}
\end{proposition}

\textbf{Remark.} The bounds in Proposition \ref{proposition:1} reveal that as $\epsilon \to 0$ (i.e., $\sigma_{\max}(\tau_{\mathrm{diffusion}}) \gg \sigma_{\max}(\tau_{\mathrm{pref}})$), the structure of the combined space is effectively "trapped" within an $\epsilon$-neighborhood of $\tau_{\mathrm{diffusion}}$. The alignment signal $\tau_{\mathrm{pref}}$ is \texttt{shadowed} by the high-variance diffusion weights, acting as negligible noise. This necessitates a scaling factor $\gamma > 1$ to amplify the signal-to-noise ratio of the alignment vector.

\begin{figure*}
    \centering
    \includegraphics[width=1\linewidth]{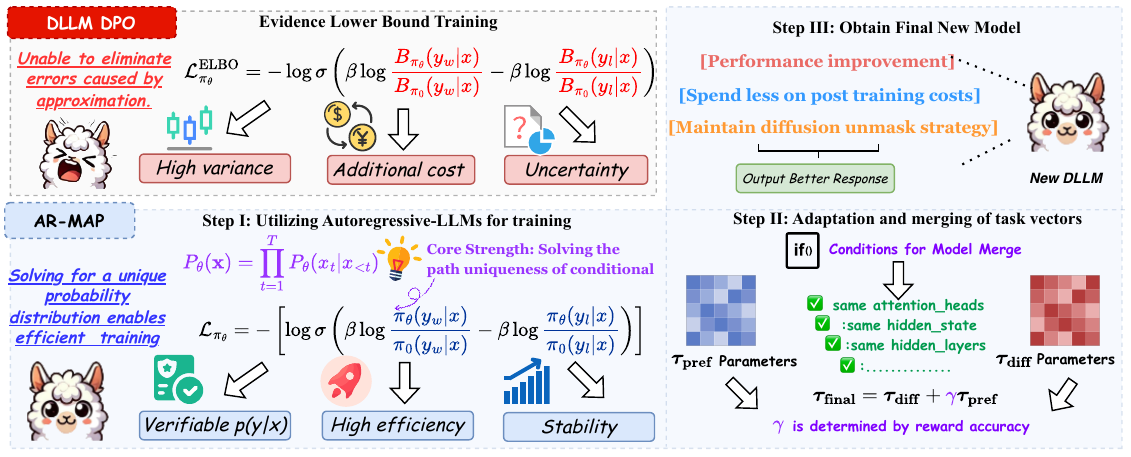}
\vspace{-1.0em}
    \caption{An overview of \ourmethod. \ourmethod has implemented a new preference alignment method through training AR LLMs and simple but effective weight transfer from AR-LLMs to DLLMs. }
    \label{fig:overview}
\end{figure*}

\subsection{RQ3: How to Effectively Scale Task Vectors}
Having established the necessity of scaling in Section \ref{section:RQ2}, we now determine the optimal granularity.

\begin{algorithm}[t]
\caption{AR-MAP Reward-based Search Strategy}
\label{alg:reward_search}
\begin{algorithmic}[1]
\REQUIRE Batch $\mathcal{B}$, Preference vector $\tau_{\text{pref}}$, Model weights $\theta_{\text{DLLM}}$
\ENSURE Optimal scaling factor $\hat{\gamma}$

\STATE Initialize $\gamma \leftarrow 1$, history set $\mathcal{S} \leftarrow \emptyset$, $Acc_{\text{best}} \leftarrow 0$

\STATE \COMMENT{Phase 1: Coarse Search }
\WHILE{True}
    \STATE \textbf{Merge:} $\theta_{\text{new}} \leftarrow \theta_{\text{DLLM}} + \gamma \cdot \tau_{\text{pref}}$
    \STATE \textbf{Evaluate:} Calculate $Acc_{\gamma}$ on $\mathcal{B}$ using Eq. (\ref{eq:reward_accuracy_search})
    \STATE \textbf{Store:} $\mathcal{S} \leftarrow \mathcal{S} \cup \{(\gamma, Acc_{\gamma})\}$
    
    \IF{$Acc_{\gamma} < Acc_{\text{best}}$}
        \STATE \textbf{break} \COMMENT{Stop if accuracy drops}
    \ENDIF
    
    \STATE $Acc_{\text{best}} \leftarrow Acc_{\gamma}$
    \STATE $\gamma \leftarrow \gamma + 2$
\ENDWHILE

\STATE \COMMENT{Phase 2: Fine-grained Check (-1)}
\STATE Evaluate $\gamma' \leftarrow \gamma - 1$, compute $Acc_{\gamma'}$, and add to $\mathcal{S}$

\STATE \textbf{return} $\hat{\gamma} = \arg\max_{(\gamma, Acc) \in \mathcal{S}} Acc$

\end{algorithmic}
\label{alg:1}
\end{algorithm}
\begin{table*}[t]
    \centering
    \renewcommand{\arraystretch}{1.0} 
    \setlength{\tabcolsep}{8pt}      
    \setlength{\aboverulesep}{3pt}   
    \setlength{\belowrulesep}{3pt}   
    \caption{Experimental Results. The best and second-best results are highlighted in \textbf{bold} and underlined, respectively. The highlighted rows represent our proposed method. The dashed line represents the segmentation of the model above after various training methods.}
    \vspace{-0.6em}
    \label{tab:results}
    \resizebox{\textwidth}{!}{
        \begin{tabular}{lccccccc}
            \toprule
            Method & GSM8K & MATH 500 & Alpacaeval & Ifeval & Arena-Hard & TruthfulQA & Avg. \\
            \midrule
            
            Qwen3-8B-Base & 85.82 & 69.60 & 55.53 & 46.76 & 57.40 & 54.41 & 61.59 \\
            \hdashline
            DPO & 87.95 \textcolor{red}{\scriptsize $\uparrow$ 2.13} & 77.80 \textcolor{red}{\scriptsize $\uparrow$ 8.20} & 63.98 \textcolor{red}{\scriptsize $\uparrow$ 8.45} & 56.83 \textcolor{red}{\scriptsize $\uparrow$ 10.07} & 60.40 \textcolor{red}{\scriptsize $\uparrow$ 3.00} & 58.69 \textcolor{red}{\scriptsize $\uparrow$ 4.28} & 67.61 \textcolor{red}{\scriptsize $\uparrow$ 6.02} \\
            \midrule
            
            SDAR-8B-Instruct & 90.29 & 72.20 & 31.30 & 59.35 & 41.08 & 48.33 & 57.09 \\
            \hdashline
            DPO & 90.43 \textcolor{red}{\scriptsize $\uparrow$ 0.14} & 72.60 \textcolor{red}{\scriptsize $\uparrow$ 0.40} & 59.50 \textcolor{red}{\scriptsize $\uparrow$ 28.20} & \underline{66.27} \textcolor{red}{\scriptsize $\uparrow$ 6.92} & 61.12 \textcolor{red}{\scriptsize $\uparrow$ 20.04} & 50.47 \textcolor{red}{\scriptsize $\uparrow$ 2.14} & 66.73 \textcolor{red}{\scriptsize $\uparrow$ 9.64} \\
            VRPO & \textbf{91.60} \textcolor{red}{\scriptsize $\uparrow$ 1.31} & \underline{73.00} \textcolor{red}{\scriptsize $\uparrow$ 0.80} & \underline{68.32} \textcolor{red}{\scriptsize $\uparrow$ 37.02} & 65.55 \textcolor{red}{\scriptsize $\uparrow$ 6.20} & \underline{68.14} \textcolor{red}{\scriptsize $\uparrow$ 27.06} & \underline{55.30} \textcolor{red}{\scriptsize $\uparrow$ 6.97} & \underline{70.32} \textcolor{red}{\scriptsize $\uparrow$ 13.23} \\
            SimPO & 90.37 \textcolor{red}{\scriptsize $\uparrow$ 0.08} & 70.80 \textcolor{green}{\scriptsize $\downarrow$ 1.40} & 51.80 \textcolor{red}{\scriptsize $\uparrow$ 20.50} & 63.79 \textcolor{red}{\scriptsize $\uparrow$ 4.44} & 61.08 \textcolor{red}{\scriptsize $\uparrow$ 20.00} & 52.32 \textcolor{red}{\scriptsize $\uparrow$ 3.99} & 65.03 \textcolor{red}{\scriptsize $\uparrow$ 7.94} \\
            \rowcolor{highlightblue} \textbf{AR-MAP} & 90.79 \textcolor{red}{\scriptsize $\uparrow$ 0.50} & \textbf{74.00} \textcolor{red}{\scriptsize $\uparrow$ 1.80} & \textbf{72.80} \textcolor{red}{\scriptsize $\uparrow$ 41.50} & \textbf{66.43} \textcolor{red}{\scriptsize $\uparrow$ 7.08} & \textbf{73.55} \textcolor{red}{\scriptsize $\uparrow$ 32.47} & \textbf{55.80} \textcolor{red}{\scriptsize $\uparrow$ 7.47} & \textbf{72.23} \textcolor{red}{\scriptsize $\uparrow$ 15.14} \\

            \midrule
            
            Qwen2.5-7B & 82.11 & 62.00 & 56.89 & 39.93 & 50.20 & 56.31 & 57.91 \\
            \hdashline
            DPO & 83.09 \textcolor{red}{\scriptsize $\uparrow$ 0.98} & 73.20 \textcolor{red}{\scriptsize $\uparrow$ 11.20} & 73.42 \textcolor{red}{\scriptsize $\uparrow$ 16.53} & 48.80 \textcolor{red}{\scriptsize $\uparrow$ 8.87} & 67.80 \textcolor{red}{\scriptsize $\uparrow$ 17.60} & 58.66 \textcolor{red}{\scriptsize $\uparrow$ 2.35} & 67.50 \textcolor{red}{\scriptsize $\uparrow$ 9.59} \\
            \midrule
            
            Dream-7B-Instruct & 79.60 & 44.40 & 52.80 & 27.94 & 55.04 & 40.12 & 49.98 \\
            \hdashline
            DPO & \underline{82.79} \textcolor{red}{\scriptsize $\uparrow$ 3.19} & 45.00 \textcolor{red}{\scriptsize $\uparrow$ 0.60} & 61.44 \textcolor{red}{\scriptsize $\uparrow$ 8.64} & 49.02 \textcolor{red}{\scriptsize $\uparrow$ 21.08} & 71.23 \textcolor{red}{\scriptsize $\uparrow$ 16.19} & 42.87 \textcolor{red}{\scriptsize $\uparrow$ 2.75} & 58.73 \textcolor{red}{\scriptsize $\uparrow$ 8.75} \\
            VRPO & \textbf{83.60} \textcolor{red}{\scriptsize $\uparrow$ 4.00} & 45.20 \textcolor{red}{\scriptsize $\uparrow$ 0.80} & \underline{69.73} \textcolor{red}{\scriptsize $\uparrow$ 16.93} & \underline{61.84} \textcolor{red}{\scriptsize $\uparrow$ 33.90} & \underline{81.80} \textcolor{red}{\scriptsize $\uparrow$ 26.76} & \underline{43.22} \textcolor{red}{\scriptsize $\uparrow$ 3.10} & \underline{64.23} \textcolor{red}{\scriptsize $\uparrow$ 14.25} \\
            SimPO & 81.90 \textcolor{red}{\scriptsize $\uparrow$ 2.30} & 44.80 \textcolor{red}{\scriptsize $\uparrow$ 0.40} & 55.63 \textcolor{red}{\scriptsize $\uparrow$ 2.83} & 41.29 \textcolor{red}{\scriptsize $\uparrow$ 13.35} & 72.20 \textcolor{red}{\scriptsize $\uparrow$ 17.16} & 41.99 \textcolor{red}{\scriptsize $\uparrow$ 1.87} & 56.30 \textcolor{red}{\scriptsize $\uparrow$ 6.32} \\
            \rowcolor{highlightblue} \textbf{AR-MAP} & 82.41 \textcolor{red}{\scriptsize $\uparrow$ 2.81} & \textbf{47.20} \textcolor{red}{\scriptsize $\uparrow$ 2.80} & \textbf{75.65} \textcolor{red}{\scriptsize $\uparrow$ 22.85} & \textbf{61.90} \textcolor{red}{\scriptsize $\uparrow$ 33.96} & \textbf{84.60} \textcolor{red}{\scriptsize $\uparrow$ 29.56} & \textbf{43.79} \textcolor{red}{\scriptsize $\uparrow$ 3.67} & \textbf{65.94} \textcolor{red}{\scriptsize $\uparrow$ 15.96} \\

            \bottomrule
        \end{tabular}
    }
\end{table*}

\textbf{AR-MAP Reward-based Search Algorithm.} 
In order to accurately find the optimal scaling factor and avoid errors caused by the non-sparsity and non-linearity proven above, we propose a reward-driven search algorithm to solve the following RQ.
\infobox{
\textbf{RQ3:} Can we regress to the DPO fitting objective to efficiently determine a global scaling factor?
}
We posit that the essence of the DPO fitting process is to steadily improve the \textbf{Batch Reward Accuracy}—defined as the consistency with which the model assigns a higher implicit reward to the preferred response $y_w$ than to the dispreferred $y_l$. 
Leveraging this property, we can determine the "implicit absorption level" of the task vector by searching for the global scaling factor that maximizes this accuracy on a sampled training batch.

Formally, let $\pi_{\gamma} \coloneqq \pi_{\theta_{\text{DLLM}} + \gamma \tau_{\text{pref}}}$ denote the policy of the DLLM merged with the preference task vector at scale $\gamma$. We define the implicit reward of this merged model as $r_\gamma(x, y) = \log \frac{\pi_{\gamma}(y | x)}{\pi_{\text{ref}}(y | x)}$. 
The optimal global scalar $\hat{\gamma}$ is determined by maximizing the pairwise discrimination capability:
\begin{equation}
\label{eq:reward_accuracy_search}
\hat{\gamma} = \arg\max_{\gamma \in [0, 1]} \frac{1}{|\mathcal{B}|} \sum_{(x, y_w, y_l) \in \mathcal{B}} \mathbb{I} \left[ r_\gamma(x, y_w) > r_\gamma(x, y_l) \right],
\end{equation}

More specifically, since DLLMs generate tokens via parallel denoising rather than sequential prediction, we assess the reward accuracy by corrupting the target response $y$ into a masked state and calculating the conditional probabilities of reconstructing these masked tokens. Formally, let $\pi_{\gamma} \coloneqq \pi_{\theta_{\text{DLLM}} + \gamma \tau_{\text{pref}}}$ denote the policy of the DLLM merged with the preference task vector at scale $\gamma$. 

We define the  reward $r_\gamma(x, y)$ of this merged model as the log-likelihood difference under the diffusion process:
\begin{equation}
\label{eq:implicit_reward}
r_\gamma(x, y) \approx \mathbb{E}_{t} \left[ \log \frac{\pi_{\gamma}(y \mid y_t = \texttt{[MASK]}, x)}{\pi_{\text{ref}}(y \mid y_t = \texttt{[MASK]}, x)} \right]
\end{equation}

where $y_t = \texttt{[MASK]}$ represents the sequence $y$ corrupted at timestep $t$ (i.e., replacing specific tokens with \texttt{[MASK]}). The optimal global scalar $\hat{\gamma}$ is then determined by maximizing pairwise discrimination capability based on this reward:
\begin{equation}
\label{eq:reward_accuracy_search}
\hat{\gamma} = \arg\max_{\gamma \in [0, 1]} \frac{1}{|\mathcal{B}|} \sum_{(x, y_w, y_l) \in \mathcal{B}} \mathbb{I} \left[ r_\gamma(x, y_w) > r_\gamma(x, y_l) \right],
\end{equation}

where $\mathcal{B}$ is a mini-batch sampled from the training set and $\mathbb{I}[\cdot]$ is the indicator function. This objective effectively filters out the destructive spectral noise by selecting the precise intensity $\gamma$ that aligns the preference knowledge with the model's ranking logic. It allows us to stably measure the fitting degree of different tasks through moderate batches, without relying on \textbf{any test set information} or \textbf{building specialized evaluation standards} for the training set. For a more specific search process and overview of \ourmethod, please refer to Algorithm \ref{alg:1} and Figure~\ref{fig:overview}.

\section{Experiments}
In this section, we conduct extensive experiments to answer the following research questions:

\begin{itemize}
[leftmargin=*,itemsep=3pt,topsep=0.pt,parsep=0pt]
    \item \textbf{RQ4:} Can \ourmethod perform excellently in different tasks?

   \item \textbf{RQ5:} If the advantages brought by AR-LLM training will manifest as differences across various task mappings.
  \item \textbf{RQ6:} What the relationship between performance changes and scaling factors?
   \item \textbf{RQ7:} Can the scaling factor search algorithm based on reward accuracy effectively find suitable coefficients?
      \item \textbf{RQ8:} Can this weight mapping be generalized to other model merging methods?
\end{itemize}

\subsection{Experiments Setup}
\textbf{Evaluation and Benchmarks.} Our evaluation covers six datasets from three domains, including \textbf{Truthfulness}: \texttt{TruthfulQA}~\citep{lin2022truthfulqa} and \texttt{Ifeval}~\citep{zhou2023instruction}. \textbf{Helpfulness}: \texttt{AlpacaEval}~\citep{li2023alpacaeval} and \texttt{Arena-Hard}~\citep{li2024crowdsourced}
. 
\textbf{Math reasoning}: \texttt{GSM8K}~\citep{cobbe2021training} and \texttt{MATH500}~\citep{hendrycks2021measuring}. For helpfulness evaluation, we use the prompt in \cite{zou2023universal}  to
evaluate the helpful win rate. For math reasoning, we test the ground truth answer in test set, respectively. For Truthfulness, we report the TruthfulQA MC2 in \texttt{TruthfulQA} and evaluate \texttt{IFeval } Inst level strict using the logic of lm-evaluation \footnote{\url{https://github.com/EleutherAI/lm-evaluation-harness}}. More specific evaluation indicators can be found in Appendix~\ref{app:mectrics}.

\textbf{Training Datasets.} Follow \cite{xu2025reward}, we conduct training using datasets corresponding to distinct preference objectives, focusing on three key aspects: helpfulness, math-reasoning, and truthfulness. For the helpfulness and truthfulness objective, we use the same data in~\cite{xu2025reward}, which selected 10K samples from \texttt{Helpsteer2}~\citep{wang2024helpsteer} and \texttt{UltraFeedback}~\citep{cui2023ultrafeedback}. For math tasks, we are training on Math training set\footnote{\url{https://huggingface.co/datasets/ankner/math-500}}.

\textbf{Baselines.} We adapt \texttt{Dream-7B-instruct} and \texttt{SDAR-8B-Instruct} as the backbone model for our experiments. For training method, we select \texttt{VRPO}~\citep{zhu2025llada} and \texttt{Simpo}~\citep{meng2024simpo} as baselines. 


\textbf{Implementation Details.} For AR-MAP Reward-based Search Strategy, we set the val-batch to 4096. We adapt LoRA adapters to achieve alignment, the LoRA rank is set to 16, and the scaling factor is set to 16. All training sets were trained on 8 ${\times}$ H20-96GB  for 3 epochs. For autoregressive training, our total batch size is 128, and for DLLMs training, the batch size is 32. For DPO rollout, our default sampling size is 8. The hyperparameter settings for the baselines are detailed in the Appendix~\ref{appendix:baselines}.

\subsection{Main Experimental Results}

\begin{table*}[t]
    \centering
    \small
    \caption{Ablation Study on Scaling Factor $\gamma$. We compare scaling factors (1-6) against our optimal factor $\hat{\gamma}$ (highlighted in gray). The models are arranged side-by-side for compact comparison.}
    \label{tab:ablation_factor_gray}
    
    \setlength{\tabcolsep}{0pt}   
    \renewcommand{\arraystretch}{1.0} 

    \begin{tabular*}{\textwidth}{@{\extracolsep{\fill}} l cccccc c c cccccc c}
        \toprule
        \multirow{3}{*}{\textbf{Dataset}} & 
        \multicolumn{7}{c}{\textbf{ SDAR-8B-Instruct}} & & 
        \multicolumn{7}{c}{\textbf{ Dream-7B-Instruct}} \\
        
        \cmidrule(lr){2-8} \cmidrule(lr){10-16}
        
         & \multicolumn{6}{c}{Scaling Factor $\gamma$} & \cellcolor{gray!10}\textbf{Ours} & 
         & \multicolumn{6}{c}{Scaling Factor $\gamma$} & \cellcolor{gray!10}\textbf{Ours} \\
         
         & 1 & 2 & 3 & 4 & 5 & 6 & \cellcolor{gray!10}($\hat{\gamma}$) & 
         & 1 & 2 & 3 & 4 & 5 & 6 & \cellcolor{gray!10}($\hat{\gamma}$) \\
        \midrule
        
        
        AlpacaEval & 52.17 & 66.67 & 70.15 & \textbf{72.80} & 64.68 & -- & \cellcolor{gray!10}\textbf{4} & & 66.71 & 72.67 & 75.16 & \textbf{75.65} & 72.95 & -- & \cellcolor{gray!10}\textbf{4} \\
        Arena-Hard & 52.91 & 65.93 & 72.34 & \textbf{73.55} & 67.20 & -- & \cellcolor{gray!10}\textbf{4} & & 72.40 & 77.00 & 83.00 & \textbf{84.60} & 80.60 & -- & \cellcolor{gray!10}\textbf{4} \\
        
        IfEval     & 61.99 & 65.35 & \textbf{66.43} & 65.71 & --    & -- & \cellcolor{gray!10}\textbf{3} & & 38.25 & 58.23 & 52.87 & \textbf{61.99} & 60.43 & -- & \cellcolor{gray!10}\textbf{4} \\
        TruthfulQA & 50.12 & 52.47 & \textbf{55.80} & 55.79 & --    & -- & \cellcolor{gray!10}\textbf{3} & & 40.98 & 41.65 & 42.37 & 43.79 & \textbf{44.57} & 44.31 & \cellcolor{gray!10}\textbf{4} \\
        
        GSM8K      & 90.44 & \textbf{90.79} & 60.37 & -- & --      & -- & \cellcolor{gray!10}\textbf{2} & & 82.10 & 82.10 & \textbf{82.41} & 82.33 & -- & -- & \cellcolor{gray!10}\textbf{3} \\
        Math500    & 72.00 & \textbf{74.00} & 60.60 & -- & --      & -- & \cellcolor{gray!10}\textbf{2} & & 45.20 & 45.40 & \textbf{47.20} & 42.80 & -- & -- & \cellcolor{gray!10}\textbf{3} \\
        
        \midrule
        \textbf{Avg.} & 63.27 & 69.20 & 64.28 & 66.96 & 65.94 & -- & \cellcolor{gray!10}\textbf{72.23} & & 57.61 & 62.84 & 63.84 & 65.19 & 64.64 & -- & \cellcolor{gray!10}\textbf{65.94} \\
        
        \bottomrule
    \end{tabular*}
\end{table*}

\textbf{[For RQ4] Obs \ding{182}: Competitive performance against training-based baselines.} Remarkably, \ourmethod not only surpasses merging baselines but also achieves performance parity with, and often exceeds, resource-intensive training methods like \texttt{VRPO}. 
For instance, on the \texttt{SDAR-8B} backbone, \ourmethod achieves an overall average score of 72.23, surpassing the 70.32 achieved by \texttt{VRPO} which requires computationally expensive sampling. 
Even on complex reasoning tasks such as \texttt{Arena-Hard}, \ourmethod maintains a decisive edge, scoring 84.60 on \texttt{Dream-7B} compared to 81.80 for \texttt{VRPO}. 
This indicates that simply transferring the ``task vector'' from an aligned AR-LLMs is sufficient to instill high-quality preference knowledge into DLLMs.

\begin{figure}
    \centering
    \includegraphics[width=1\linewidth]{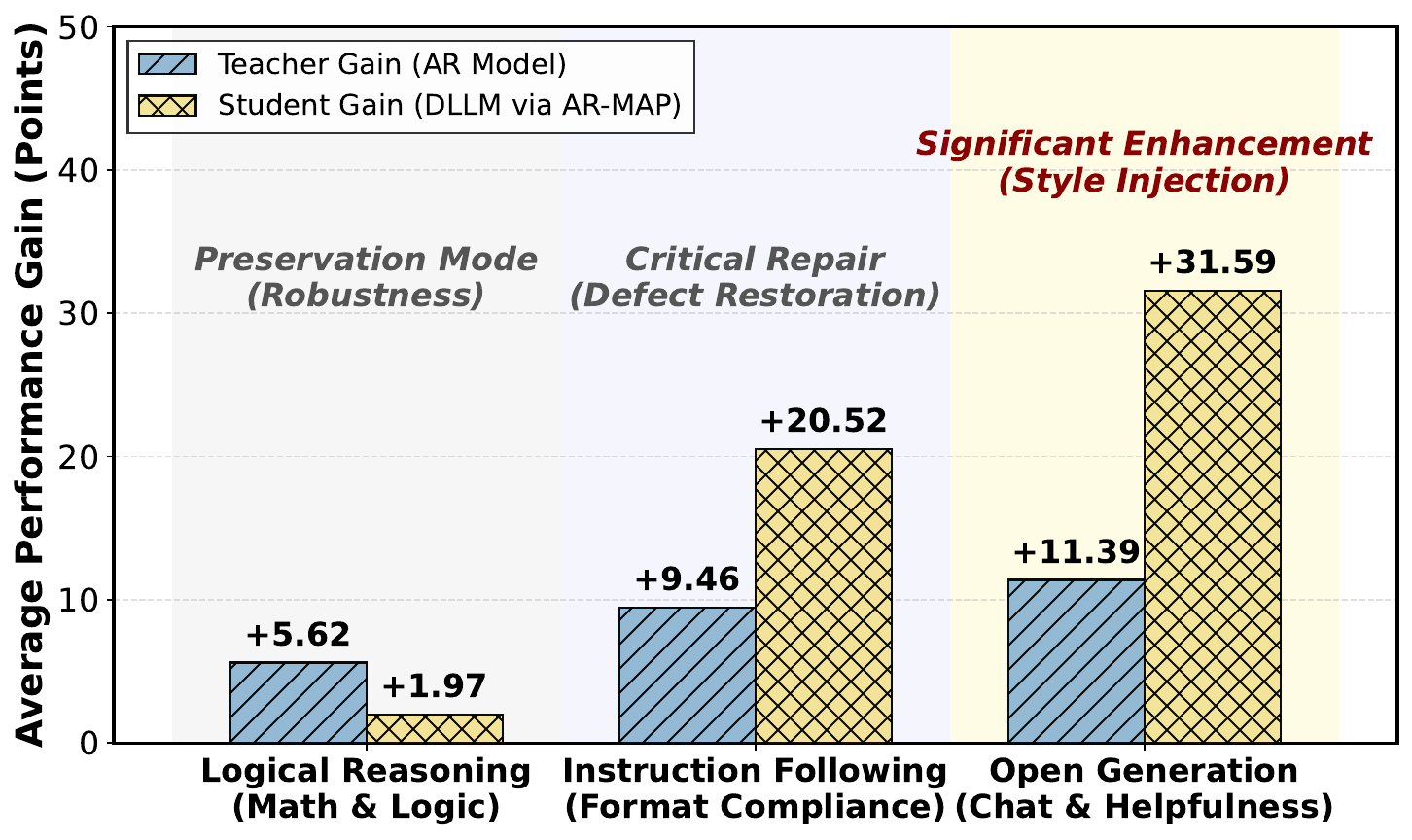}
    \caption{\textbf{Analysis of Domain-Specific Alignment Transfer.} We visualize the performance gains of the AR teacher compared to the DLLM student across distinct domains. The results highlight that \ourmethod adapts its transfer mechanism.}
    \label{fig:lingyu}
    \vspace{-1.6em}
\end{figure}

\begin{figure}[h]
    \centering
    \includegraphics[width=0.9\linewidth]{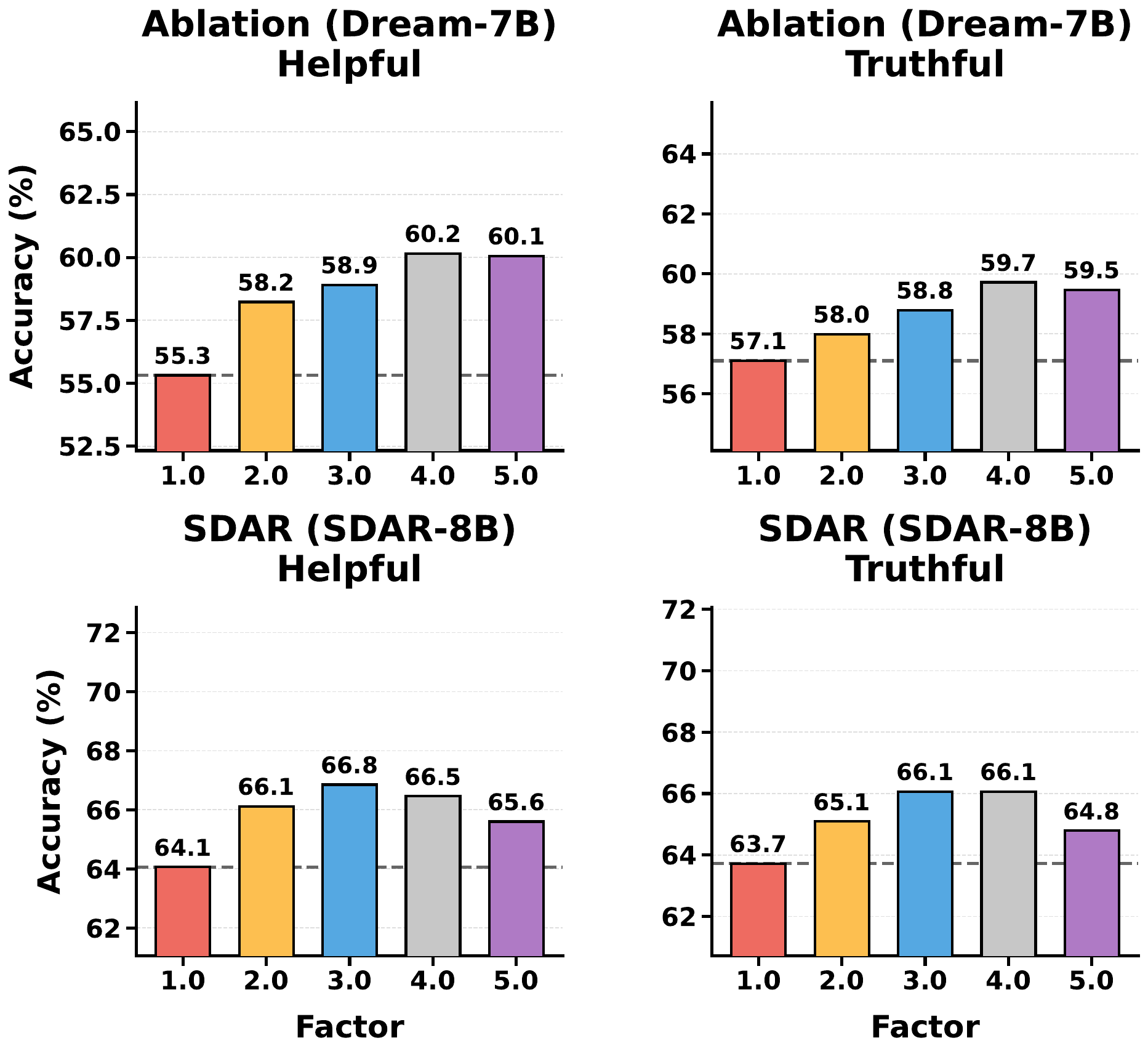}
    \vspace{-0.6em}
    \caption{Examples of different models that perform reward accuracy calculation on 4096 training set samples.}
    \label{fig:accuracy-case}
    \vspace{-1.6em}
\end{figure}

\textbf{[For RQ5] Obs \ding{183}: Mapping laws exhibit significant domain heterogeneity.} As shown in Figure~\ref{fig:lingyu}, our analysis reveals that the alignment transfer is not uniform but adapts to the student's intrinsic deficits. Specifically, \ourmethod operates in a \textit{Preservation Mode} for logical reasoning, yielding conservative gains (+1.97) compared to the teacher (+5.62) to maintain robust priors. Conversely, it shifts to a \textit{Critical Repair} mechanism for instruction following (+20.52 vs +9.47) to restore structural constraints, and acts as a \textit{Significant Enhancement} engine in open-ended generation (+31.59), significantly amplifying the teacher's signal (+11.39) to fill the preference void in diffusion models.

\subsection{Ablation Study}
In this Section, we observe the weight mapping relationship between DLLMs and AR-LLMs by gradually increasing the weight factor.

\textbf{[For RQ6] Obs \ding{184}: Relationship between performance and scaling factor.} 
The ablation study in Table~\ref{tab:ablation_factor_gray} reveals a critical tension between scaling magnitude $\gamma$ and task-specific resilience. 
We observe a sharp dichotomy: reasoning-intensive domains (e.g., \texttt{GSM8K}) exhibit structural fragility, suffering catastrophic degradation from over-scaling (notably, \texttt{SDAR} collapses at $\gamma > 2$), whereas open-ended tasks (e.g., \texttt{AlpacaEval}) necessitate aggressive parameter injection ($\gamma \approx 4$) to effectuate stylistic alignment. 
This divergence renders uniform scaling strategies suboptimal, validating the architectural necessity of our dynamic reward-based search to precisely locate the equilibrium between capability absorption and logical integrity.

\textbf{[For RQ7] Obs \ding{185}: Can the scaling factor search algorithm based on reward accuracy effectively find suitable coefficients?} As shown in Table~\ref{tab:ablation_factor_gray} and Figure~\ref{fig:accuracy-case}, our reward-based search strategy demonstrates exceptional precision and adaptability, accurately pinpointing the optimal scaling factor $\hat{\gamma}$ in \textbf{11 out of 12} experimental settings. 
Critically, the algorithm successfully captures the nuanced sensitivity differences between models and tasks: for instance, in math reasoning (e.g., Math500), it correctly identifies that SDAR requires a conservative scaling ($\hat{\gamma}=2$) to prevent performance collapse, whereas Dream tolerates a higher intensity ($\hat{\gamma}=3$); conversely, for open-ended generation (e.g., AlpacaEval), it adaptively pushes the factor to $\hat{\gamma}=4$ for both models to maximize helpfulness gains. 
Although a marginal deviation exists in the \textit{Dream-7B/TruthfulQA} task (selecting $\gamma=4$ vs. optimal $\gamma=5$), the chosen factor still yields near-optimal performance ($\Delta < 0.8\%$), confirming that maximizing Batch Reward Accuracy serves as a robust, model-aware proxy that safely navigates the trade-off between alignment transfer and knowledge retention.

\begin{table*}[t]
    \centering
    \small
    \renewcommand{\arraystretch}{1.0} 
    \setlength{\tabcolsep}{8pt}       
    \setlength{\aboverulesep}{3pt}    
    \setlength{\belowrulesep}{3pt}    

    \caption{Ablation Study on Scaling Factor with Ties and Dare + AR-MAP. We investigate the performance impact of varying $\gamma$ from 2.0 to 7.0 for Ties and 1.0 to 2.0 for Dare.}
    \label{tab:ties_armap_ablation}
    \vspace{-0.1em}    
    \resizebox{\textwidth}{!}{
        \begin{tabular}{l cccccc}
            \toprule
            Method & MATH 500 & AlpacaEval & IfEval & Arena-Hard & TruthfulQA & Avg. \\
            \midrule
            
            Ties + AR-MAP ($\gamma=2.0$) & \underline{70.40} & 63.98 & \underline{64.99} & 62.36 & 51.28 & \underline{62.60} \\

            Ties + AR-MAP ($\gamma=3.0$) & 64.80 & 64.60 & 64.87 & 65.66 & 52.42 & 62.47 \\
            
            Ties + AR-MAP ($\gamma=4.0$) & 55.00 & \textbf{72.05} & \textbf{66.19} & \textbf{68.74} & 53.24 & \textbf{63.04} \\
            
            Ties + AR-MAP ($\gamma=5.0$) & 37.80 & \underline{69.69} & 61.55 & \underline{66.13} & 53.92 & 57.82 \\
            
            Ties + AR-MAP ($\gamma=6.0$) & 15.40 & 68.82 & 61.63 & 63.73 & \underline{54.52} & 52.82 \\
            
            Ties + AR-MAP ($\gamma=7.0$) & 7.00 & 60.75 & 61.99 & 59.84 & \textbf{55.07} & 48.93 \\
            
            DARE + AR-MAP ($\gamma=1.0$) & \textbf{71.20} & 51.34 & 61.63 & 58.84 & 49.97 & 58.60 \\
            
            DARE + AR-MAP ($\gamma=2.0$) & 19.03 & 0.87 & 26.74 & 0.40 & 51.29 & 19.66 \\
            \bottomrule
        \end{tabular}
    }
    \vspace{-0.5em}
\end{table*}
\begin{figure*}[t]
    \centering
    \includegraphics[width=1\linewidth]{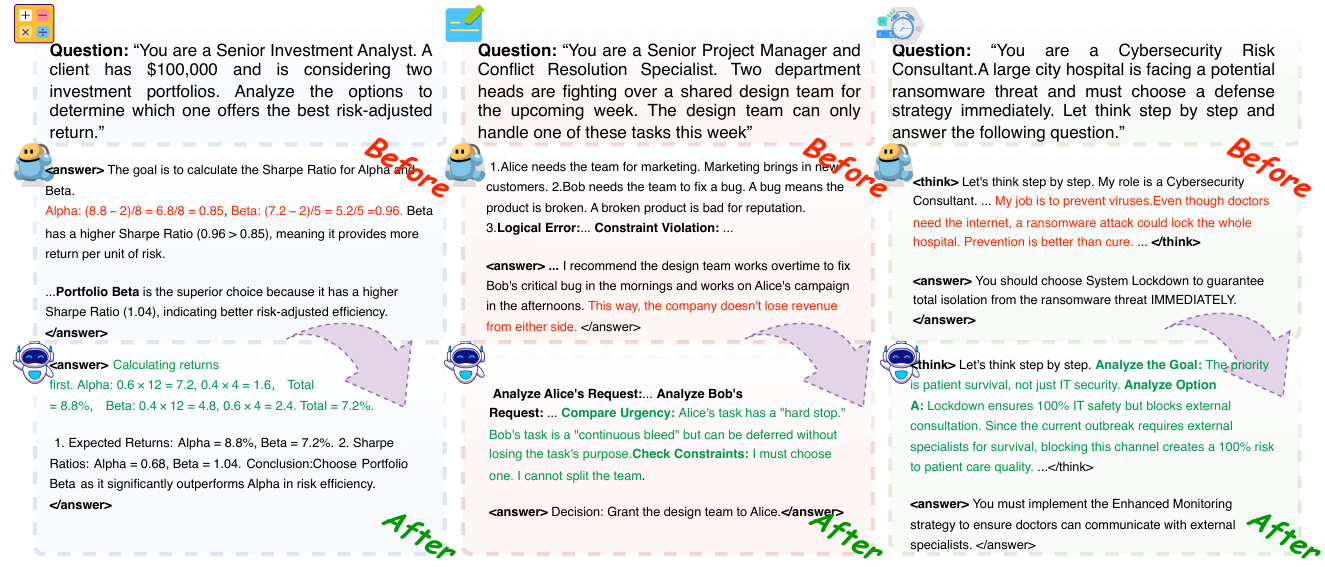}
   \caption{Examples of changes in model performance on different tasks before and after \ourmethod.}
    \label{fig:placeholder}
    \vspace{-1.0em}
\end{figure*}

\subsection{Generalization Experiment}
In this section, we are trying to investigate whether other merge methods can also achieve migration through scaling up.
We use \texttt{Ties}~\citep{yadav2023ties} and \texttt{Dare}~\citep{yu2024language} as the test objects and obtain the following conclusions, please refer to the Appendix~\ref{appendix:baselines} for specific hyperparameter configurations.

\textbf{[For RQ8] Obs \ding{186}: AR LLMs training, DLLMs absorbing this paradigm can be generalized to other model merging methods.} As shown in Figure~\ref{fig:placeholder} and Table~\ref{tab:ties_armap_ablation}, for \texttt{Ties}, performance follows an inverted U-shaped trajectory, peaking at an overall average score of 63.04 with $\gamma = 4.0$, which outperforms both the conservative ($\gamma = 2.0$) and over-scaled ($\gamma = 7.0$) configurations. Furthermore, \texttt{DARE} exhibits extreme structural fragility, suffering a catastrophic performance collapse at just $\gamma = 2.0$, indicating that sensitivity to weight scaling is highly method-dependent.  Ultimately, these results confirm that \ourmethod 's findings are reasonable and generalizable.
\section{Related Work}
\paragraph{Diffusion Large Language Models} (DLLMs) leverage the principles of denoising diffusion to model language understanding and generation~\cite{diffubert, ld4lg,wang2026everything}. Research primarily follows two distinct paths~\cite{dllm_survey}: continuous denoising models, which perform denoising within the continuous embedding space after mapping tokens~\cite{diffuseq, ssdlm}, and Masked Diffusion Models, which operate via a mask-remask mechanism directly~\cite{diffullama, ddpo, nie2025large}. Recent studies demonstrate that DLLMs achieve performance comparable to Autoregressive (AR) LLMs~\cite{dllm4se, gemini_diffusion, diffuguard, bie2025llada2}. Works like LLaDA~\cite{nie2025large} and Seed Diffusion~\cite{seed_diffusion} train models from scratch using randomly initialized weights~\cite{smdm, block_diffusion, lladamoe}. Conversely, works such as Dream~\cite{ye2025dream}, DiffuCoder~\cite{diffucoder}, and SDAR~\cite{cheng2025sdar} utilize AR LLM weights as a starting point for continued pre-training, aiming to simplify the training process~\cite{fastdllmv2, wang2025revolutionizing, wedlm, stable_diffucoder}.

\paragraph{Model Merging.} Model merging~\citep{yang2024model,cheng2025whoever,akiba2025evolutionary} has emerged as a promising approach to enhance model capabilities without the need for access to raw training data or extensive computational resources, offering a cost-effective way to boost the performance of large language models (LLMs). Model merging methods generally fall into two types: pre-merging~\citep{ilharco2022editing,wortsman2022model,stoica2023zipit,zhang2024knowledge} approaches, which align weights or architectures before fusion, and during-merging strategies~\citep{yu2024language,yang2024representation}, which combine models using techniques like averaging, weighting, or routing to resolve task conflicts. Previous methods focus on model merging in the field of autoregression or across modalities~\citep{hu2025can}, while \ourmethod attempt to uncover the patterns of model merging between different regression models.

\section{Conclusion}

We explore the latent weight transferability between the divergent paradigms of AR-LLMs and DLLMs. 
We find that alignment knowledge from AR models must be significantly scaled to effectively penetrate the high-variance parameter space of diffusion models. 
Leveraging this discovery, our proposed \ourmethod framework enables DLLMs to efficiently inherit preference alignment capabilities, bridging the gap between autoregressive and diffusion architectures. 
Extensive experiments demonstrate that \ourmethod performs excellently on different datasets.

\bibliography{example_paper}

@article{xiong2024autoregressive,
  title={Autoregressive models in vision: A survey},
  author={Xiong, Jing and Liu, Gongye and Huang, Lun and Wu, Chengyue and Wu, Taiqiang and Mu, Yao and Yao, Yuan and Shen, Hui and Wan, Zhongwei and Huang, Jinfa and others},
  journal={arXiv preprint arXiv:2411.05902},
  year={2024}
}

@article{wang2025d2,
  title={d2: Improved techniques for training reasoning diffusion language models},
  author={Wang, Guanghan and Schiff, Yair and Turok, Gilad and Kuleshov, Volodymyr},
  journal={arXiv preprint arXiv:2509.21474},
  year={2025}
}

@article{xie2025step,
  title={Step-Aware Policy Optimization for Reasoning in Diffusion Large Language Models},
  author={Xie, Shaoan and Kong, Lingjing and Song, Xiangchen and Dong, Xinshuai and Chen, Guangyi and Xing, Eric P and Zhang, Kun},
  journal={arXiv preprint arXiv:2510.01544},
  year={2025}
}

@inproceedings{deng2025following,
  title={Following the autoregressive nature of llm embeddings via compression and alignment},
  author={Deng, Jingcheng and Jiang, Zhongtao and Pang, Liang and Wei, Zihao and Chen, Liwei and Xu, Kun and Song, Yang and Shen, Huawei and Cheng, Xueqi},
  booktitle={Proceedings of the 2025 Conference on Empirical Methods in Natural Language Processing},
  pages={12672--12688},
  year={2025}
}

@article{wang2025revolutionizing,
  title={Revolutionizing reinforcement learning framework for diffusion large language models},
  author={Wang, Yinjie and Yang, Ling and Li, Bowen and Tian, Ye and Shen, Ke and Wang, Mengdi},
  journal={arXiv preprint arXiv:2509.06949},
  year={2025}
}

@article{wu2025fast,
  title={Fast-dllm: Training-free acceleration of diffusion llm by enabling kv cache and parallel decoding},
  author={Wu, Chengyue and Zhang, Hao and Xue, Shuchen and Liu, Zhijian and Diao, Shizhe and Zhu, Ligeng and Luo, Ping and Han, Song and Xie, Enze},
  journal={arXiv preprint arXiv:2505.22618},
  year={2025}
}

@article{song2025sparse,
  title={Sparse-dllm: Accelerating diffusion llms with dynamic cache eviction},
  author={Song, Yuerong and Liu, Xiaoran and Li, Ruixiao and Liu, Zhigeng and Huang, Zengfeng and Guo, Qipeng and He, Ziwei and Qiu, Xipeng},
  journal={arXiv preprint arXiv:2508.02558},
  year={2025}
}

@article{ziegler2019fine,
  title={Fine-tuning language models from human preferences},
  author={Ziegler, Daniel M and Stiennon, Nisan and Wu, Jeffrey and Brown, Tom B and Radford, Alec and Amodei, Dario and Christiano, Paul and Irving, Geoffrey},
  journal={arXiv preprint arXiv:1909.08593},
  year={2019}
}

@article{ji2026thinking,
  title={Thinking with Map: Reinforced Parallel Map-Augmented Agent for Geolocalization},
  author={Ji, Yuxiang and Wang, Yong and Ma, Ziyu and Hu, Yiming and Huang, Hailang and Hu, Xuecai and Chen, Guanhua and Wu, Liaoni and Chu, Xiangxiang},
  journal={arXiv preprint arXiv:2601.05432},
  year={2026}
}

@article{wang2024helpsteer,
  title={Helpsteer 2: Open-source dataset for training top-performing reward models},
  author={Wang, Zhilin and Dong, Yi and Delalleau, Olivier and Zeng, Jiaqi and Shen, Gerald and Egert, Daniel and Zhang, Jimmy and Sreedhar, Makesh Narsimhan and Kuchaiev, Oleksii},
  journal={Advances in Neural Information Processing Systems},
  volume={37},
  pages={1474--1501},
  year={2024}
}

@article{stewart1993early,
  title={On the early history of the singular value decomposition},
  author={Stewart, Gilbert W},
  journal={SIAM review},
  volume={35},
  number={4},
  pages={551--566},
  year={1993},
  publisher={SIAM}
}

@article{wang2025comprehensive,
  title={A comprehensive survey in llm (-agent) full stack safety: Data, training and deployment},
  author={Wang, Kun and Zhang, Guibin and Zhou, Zhenhong and Wu, Jiahao and Yu, Miao and Zhao, Shiqian and Yin, Chenlong and Fu, Jinhu and Yan, Yibo and Luo, Hanjun and others},
  journal={arXiv preprint arXiv:2504.15585},
  year={2025}
}

@article{ilharco2022editing,
  title={Editing models with task arithmetic},
  author={Ilharco, Gabriel and Ribeiro, Marco Tulio and Wortsman, Mitchell and Gururangan, Suchin and Schmidt, Ludwig and Hajishirzi, Hannaneh and Farhadi, Ali},
  journal={arXiv preprint arXiv:2212.04089},
  year={2022}
}

@inproceedings{sun2025task,
  title={Task arithmetic in trust region: A training-free model merging approach to navigate knowledge conflicts},
  author={Sun, Wenju and Li, Qingyong and Wang, Wen and Geng, Yangliao and Li, Boyang},
  booktitle={Proceedings of the 33rd ACM International Conference on Multimedia},
  pages={5178--5187},
  year={2025}
}

@article{akiba2025evolutionary,
  title={Evolutionary optimization of model merging recipes},
  author={Akiba, Takuya and Shing, Makoto and Tang, Yujin and Sun, Qi and Ha, David},
  journal={Nature Machine Intelligence},
  volume={7},
  number={2},
  pages={195--204},
  year={2025},
  publisher={Nature Publishing Group UK London}
}

@article{li2025adacurl,
  title={Adacurl: Adaptive curriculum reinforcement learning with invalid sample mitigation and historical revisiting},
  author={Li, Renda and Huang, Hailang and Wei, Fei and Xiong, Feng and Wang, Yong and Chu, Xiangxiang},
  journal={arXiv preprint arXiv:2511.09478},
  year={2025}
}

@article{wang2026everything,
  title={Everything in Its Place: Benchmarking Spatial Intelligence of Text-to-Image Models},
  author={Wang, Zengbin and Hu, Xuecai and Wang, Yong and Xiong, Feng and Zhang, Man and Chu, Xiangxiang},
  journal={arXiv preprint arXiv:2601.20354},
  year={2026}
}

@article{chu2025gpg,
  title={Gpg: A simple and strong reinforcement learning baseline for model reasoning},
  author={Chu, Xiangxiang and Huang, Hailang and Zhang, Xiao and Wei, Fei and Wang, Yong},
  journal={arXiv preprint arXiv:2504.02546},
  year={2025}
}

@article{ma2025and,
  title={Where and What Matters: Sensitivity-Aware Task Vectors for Many-Shot Multimodal In-Context Learning},
  author={Ma, Ziyu and Gou, Chenhui and Hu, Yiming and Wang, Yong and Chu, Xiangxiang and Zhuang, Bohan and Cai, Jianfei},
  journal={arXiv preprint arXiv:2511.08246},
  year={2025}
}

@article{xiong2025hs,
  title={HS-STAR: Hierarchical Sampling for Self-Taught Reasoners via Difficulty Estimation and Budget Reallocation},
  author={Xiong, Feng and Xu, Hongling and Wang, Yifei and Cheng, Runxi and Wang, Yong and Chu, Xiangxiang},
  journal={arXiv preprint arXiv:2505.19866},
  year={2025}
}

@article{ji2025tree,
  title={Tree search for llm agent reinforcement learning},
  author={Ji, Yuxiang and Ma, Ziyu and Wang, Yong and Chen, Guanhua and Chu, Xiangxiang and Wu, Liaoni},
  journal={arXiv preprint arXiv:2509.21240},
  year={2025}
}

@article{stoica2023zipit,
  title={Zipit! merging models from different tasks without training},
  author={Stoica, George and Bolya, Daniel and Bjorner, Jakob and Ramesh, Pratik and Hearn, Taylor and Hoffman, Judy},
  journal={arXiv preprint arXiv:2305.03053},
  year={2023}
}

@article{yang2024representation,
  title={Representation surgery for multi-task model merging},
  author={Yang, Enneng and Shen, Li and Wang, Zhenyi and Guo, Guibing and Chen, Xiaojun and Wang, Xingwei and Tao, Dacheng},
  journal={arXiv preprint arXiv:2402.02705},
  year={2024}
}

@article{zhang2024knowledge,
  title={Knowledge composition using task vectors with learned anisotropic scaling},
  author={Zhang, Frederic Z and Albert, Paul and Rodriguez-Opazo, Cristian and van den Hengel, Anton and Abbasnejad, Ehsan},
  journal={Advances in Neural Information Processing Systems},
  volume={37},
  pages={67319--67354},
  year={2024}
}

@inproceedings{wortsman2022model,
  title={Model soups: averaging weights of multiple fine-tuned models improves accuracy without increasing inference time},
  author={Wortsman, Mitchell and Ilharco, Gabriel and Gadre, Samir Ya and Roelofs, Rebecca and Gontijo-Lopes, Raphael and Morcos, Ari S and Namkoong, Hongseok and Farhadi, Ali and Carmon, Yair and Kornblith, Simon and others},
  booktitle={International conference on machine learning},
  pages={23965--23998},
  year={2022},
  organization={PMLR}
}

@article{cheng2025whoever,
  title={Whoever started the interference should end it: Guiding data-free model merging via task vectors},
  author={Cheng, Runxi and Xiong, Feng and Wei, Yongxian and Zhu, Wanyun and Yuan, Chun},
  journal={arXiv preprint arXiv:2503.08099},
  year={2025}
}

@article{team2024qwen2,
  title={Qwen2 technical report},
  author={Team, Qwen and others},
  journal={arXiv preprint arXiv:2407.10671},
  volume={2},
  number={3},
  year={2024}
}

@article{yang2025qwen3,
  title={Qwen3 technical report},
  author={Yang, An and Li, Anfeng and Yang, Baosong and Zhang, Beichen and Hui, Binyuan and Zheng, Bo and Yu, Bowen and Gao, Chang and Huang, Chengen and Lv, Chenxu and others},
  journal={arXiv preprint arXiv:2505.09388},
  year={2025}
}

@article{team2025every,
  title={Every Activation Boosted: Scaling General Reasoner to 1 Trillion Open Language Foundation},
  author={Team, Ling and Li, Ang and Liu, Ben and Hu, Binbin and Li, Bing and Zeng, Bingwei and Ye, Borui and Tang, Caizhi and Tian, Changxin and Huang, Chao and others},
  journal={arXiv preprint arXiv:2510.22115},
  year={2025}
}

@article{lee2025instructpatentgpt,
  title={InstructPatentGPT: training patent language models to follow instructions with human feedback},
  author={Lee, Jieh-Sheng},
  journal={Artificial Intelligence and Law},
  volume={33},
  number={3},
  pages={739--782},
  year={2025},
  publisher={Springer}
}

@article{huang2025pc,
  title={Pc-sampler: Position-aware calibration of decoding bias in masked diffusion models},
  author={Huang, Pengcheng and Liu, Shuhao and Liu, Zhenghao and Yan, Yukun and Wang, Shuo and Chen, Zulong and Xiao, Tong},
  journal={arXiv preprint arXiv:2508.13021},
  year={2025}
}

@article{cui2023ultrafeedback,
  title={Ultrafeedback: Boosting language models with high-quality feedback},
  author={Cui, Ganqu and Yuan, Lifan and Ding, Ning and Yao, Guanming and Zhu, Wei and Ni, Yuan and Xie, Guotong and Liu, Zhiyuan and Sun, Maosong},
  year={2023}
}

@inproceedings{lin2022truthfulqa,
  title={Truthfulqa: Measuring how models mimic human falsehoods},
  author={Lin, Stephanie and Hilton, Jacob and Evans, Owain},
  booktitle={Proceedings of the 60th annual meeting of the association for computational linguistics (volume 1: long papers)},
  pages={3214--3252},
  year={2022}
}

@article{zhou2023instruction,
  title={Instruction-following evaluation for large language models},
  author={Zhou, Jeffrey and Lu, Tianjian and Mishra, Swaroop and Brahma, Siddhartha and Basu, Sujoy and Luan, Yi and Zhou, Denny and Hou, Le},
  journal={arXiv preprint arXiv:2311.07911},
  year={2023}
}

@article{xu2025reward,
  title={REWARD CONSISTENCY: Improving Multi-Objective Alignment from a Data-Centric Perspective},
  author={Xu, Zhihao and Tong, Yongqi and Zhang, Xin and Zhou, Jun and Wang, Xiting},
  journal={arXiv preprint arXiv:2504.11337},
  year={2025}
}

@article{huang2025reinforcing,
  title={Reinforcing the diffusion chain of lateral thought with diffusion language models},
  author={Huang, Zemin and Chen, Zhiyang and Wang, Zijun and Li, Tiancheng and Qi, Guo-Jun},
  journal={arXiv preprint arXiv:2505.10446},
  year={2025}
}

@misc{li2023alpacaeval,
  title={Alpacaeval: An automatic evaluator of instruction-following models},
  author={Li, Xuechen and Zhang, Tianyi and Dubois, Yann and Taori, Rohan and Gulrajani, Ishaan and Guestrin, Carlos and Liang, Percy and Hashimoto, Tatsunori B},
  year={2023}
}

@article{cobbe2021training,
  title={Training verifiers to solve math word problems},
  author={Cobbe, Karl and Kosaraju, Vineet and Bavarian, Mohammad and Chen, Mark and Jun, Heewoo and Kaiser, Lukasz and Plappert, Matthias and Tworek, Jerry and Hilton, Jacob and Nakano, Reiichiro and others},
  journal={arXiv preprint arXiv:2110.14168},
  year={2021}
}

@article{meng2024simpo,
  title={Simpo: Simple preference optimization with a reference-free reward},
  author={Meng, Yu and Xia, Mengzhou and Chen, Danqi},
  journal={Advances in Neural Information Processing Systems},
  volume={37},
  pages={124198--124235},
  year={2024}
}

@article{yang2024model,
  title={Model Merging in LLMs, MLLMs, and Beyond: Methods, Theories, Applications, and Opportunities},
  author={Yang, Enneng and Shen, Li and Guo, Guibing and Wang, Xingwei and Cao, Xiaochun and Zhang, Jie and Tao, Dacheng},
  journal={ACM Computing Surveys},
  year={2024},
  publisher={ACM New York, NY}
}

@article{zou2023universal,
  title={Universal and transferable adversarial attacks on aligned language models},
  author={Zou, Andy and Wang, Zifan and Carlini, Nicholas and Nasr, Milad and Kolter, J Zico and Fredrikson, Matt},
  journal={arXiv preprint arXiv:2307.15043},
  year={2023}
}

@article{hendrycks2021measuring,
  title={Measuring mathematical problem solving with the math dataset},
  author={Hendrycks, Dan and Burns, Collin and Kadavath, Saurav and Arora, Akul and Basart, Steven and Tang, Eric and Song, Dawn and Steinhardt, Jacob},
  journal={arXiv preprint arXiv:2103.03874},
  year={2021}
}

@article{li2024crowdsourced,
  title={From crowdsourced data to high-quality benchmarks: Arena-hard and benchbuilder pipeline},
  author={Li, Tianle and Chiang, Wei-Lin and Frick, Evan and Dunlap, Lisa and Wu, Tianhao and Zhu, Banghua and Gonzalez, Joseph E and Stoica, Ion},
  journal={arXiv preprint arXiv:2406.11939},
  year={2024}
}

@article{tang2025wd1,
  title={wd1: Weighted policy optimization for reasoning in diffusion language models},
  author={Tang, Xiaohang and Dolga, Rares and Yoon, Sangwoong and Bogunovic, Ilija},
  journal={arXiv preprint arXiv:2507.08838},
  year={2025}
}

@article{yu2025dapo,
  title={Dapo: An open-source llm reinforcement learning system at scale},
  author={Yu, Qiying and Zhang, Zheng and Zhu, Ruofei and Yuan, Yufeng and Zuo, Xiaochen and Yue, Yu and Dai, Weinan and Fan, Tiantian and Liu, Gaohong and Liu, Lingjun and others},
  journal={arXiv preprint arXiv:2503.14476},
  year={2025}
}

@article{yadav2023ties,
  title={Ties-merging: Resolving interference when merging models},
  author={Yadav, Prateek and Tam, Derek and Choshen, Leshem and Raffel, Colin A and Bansal, Mohit},
  journal={Advances in Neural Information Processing Systems},
  volume={36},
  pages={7093--7115},
  year={2023}
}

@article{hu2025can,
  title={Can MLLMs Absorb Math Reasoning Abilities from LLMs as Free Lunch?},
  author={Hu, Yijie and Zhou, Zihao and Huang, Kaizhu and Huang, Xiaowei and Wang, Qiufeng},
  journal={arXiv preprint arXiv:2510.14387},
  year={2025}
}

@inproceedings{yu2024language,
  title={Language models are super mario: Absorbing abilities from homologous models as a free lunch},
  author={Yu, Le and Yu, Bowen and Yu, Haiyang and Huang, Fei and Li, Yongbin},
  booktitle={Forty-first International Conference on Machine Learning},
  year={2024}
}

@article{bie2025llada2,
  title={Llada2. 0: Scaling up diffusion language models to 100b},
  author={Bie, Tiwei and Cao, Maosong and Chen, Kun and Du, Lun and Gong, Mingliang and Gong, Zhuochen and Gu, Yanmei and Hu, Jiaqi and Huang, Zenan and Lan, Zhenzhong and others},
  journal={arXiv preprint arXiv:2512.15745},
  year={2025}
}

@article{ye2025dream,
  title={Dream 7b: Diffusion large language models},
  author={Ye, Jiacheng and Xie, Zhihui and Zheng, Lin and Gao, Jiahui and Wu, Zirui and Jiang, Xin and Li, Zhenguo and Kong, Lingpeng},
  journal={arXiv preprint arXiv:2508.15487},
  year={2025}
}

@article{xie2025dream,
  title={Dream-coder 7b: An open diffusion language model for code},
  author={Xie, Zhihui and Ye, Jiacheng and Zheng, Lin and Gao, Jiahui and Dong, Jingwei and Wu, Zirui and Zhao, Xueliang and Gong, Shansan and Jiang, Xin and Li, Zhenguo and others},
  journal={arXiv preprint arXiv:2509.01142},
  year={2025}
}

@article{grattafiori2024llama,
  title={The llama 3 herd of models},
  author={Grattafiori, Aaron and Dubey, Abhimanyu and Jauhri, Abhinav and Pandey, Abhinav and Kadian, Abhishek and Al-Dahle, Ahmad and Letman, Aiesha and Mathur, Akhil and Schelten, Alan and Vaughan, Alex and others},
  journal={arXiv preprint arXiv:2407.21783},
  year={2024}
}

@article{ahmed2025qwen,
  title={Qwen 2.5: A comprehensive review of the leading resource-efficient llm with potentioal to surpass all competitors},
  author={Ahmed, Imtiaz and Islam, Sadman and Datta, Partha Protim and Kabir, Imran and Chowdhury, Naseef Ur Rahman and Haque, Ahshanul},
  journal={Authorea Preprints},
  year={2025},
  publisher={Authorea}
}

@article{zhu2025llada,
  title={LLaDA 1.5: Variance-Reduced Preference Optimization for Large Language Diffusion Models},
  author={Zhu, Fengqi and Wang, Rongzhen and Nie, Shen and Zhang, Xiaolu and Wu, Chunwei and Hu, Jun and Zhou, Jun and Chen, Jianfei and Lin, Yankai and Wen, Ji-Rong and others},
  journal={arXiv preprint arXiv:2505.19223},
  year={2025}
}

@article{rafailov2023direct,
  title={Direct preference optimization: Your language model is secretly a reward model},
  author={Rafailov, Rafael and Sharma, Archit and Mitchell, Eric and Manning, Christopher D and Ermon, Stefano and Finn, Chelsea},
  journal={Advances in neural information processing systems},
  volume={36},
  pages={53728--53741},
  year={2023}
}

@article{ye2023diffusion,
  title={Diffusion language models can perform many tasks with scaling and instruction-finetuning},
  author={Ye, Jiasheng and Zheng, Zaixiang and Bao, Yu and Qian, Lihua and Gu, Quanquan},
  journal={arXiv preprint arXiv:2308.12219},
  year={2023}
}

@article{zhang2025survey,
  title={A survey on parallel text generation: From parallel decoding to diffusion language models},
  author={Zhang, Lingzhe and Fang, Liancheng and Duan, Chiming and He, Minghua and Pan, Leyi and Xiao, Pei and Huang, Shiyu and Zhai, Yunpeng and Hu, Xuming and Yu, Philip S and others},
  journal={arXiv preprint arXiv:2508.08712},
  year={2025}
}

@article{nie2025large,
  title={Large language diffusion models},
  author={Nie, Shen and Zhu, Fengqi and You, Zebin and Zhang, Xiaolu and Ou, Jingyang and Hu, Jun and Zhou, Jun and Lin, Yankai and Wen, Ji-Rong and Li, Chongxuan},
  journal={arXiv preprint arXiv:2502.09992},
  year={2025}
}

@article{cheng2025sdar,
  title={Sdar: A synergistic diffusion-autoregression paradigm for scalable sequence generation},
  author={Cheng, Shuang and Bian, Yihan and Liu, Dawei and Zhang, Linfeng and Yao, Qian and Tian, Zhongbo and Wang, Wenhai and Guo, Qipeng and Chen, Kai and Qi, Biqing and others},
  journal={arXiv preprint arXiv:2510.06303},
  year={2025}
}

@article{nie2024scaling,
  title={Scaling up masked diffusion models on text},
  author={Nie, Shen and Zhu, Fengqi and Du, Chao and Pang, Tianyu and Liu, Qian and Zeng, Guangtao and Lin, Min and Li, Chongxuan},
  journal={arXiv preprint arXiv:2410.18514},
  year={2024}
}

@misc{dllm_survey,
      title={A Survey on Diffusion Language Models}, 
      author={Tianyi Li and Mingda Chen and Bowei Guo and Zhiqiang Shen},
      year={2025},
      eprint={2508.10875},
      archivePrefix={arXiv},
      primaryClass={cs.CL},
      url={https://arxiv.org/abs/2508.10875}, 
}

@misc{diffuseq,
      title={DiffuSeq: Sequence to Sequence Text Generation with Diffusion Models}, 
      author={Shansan Gong and Mukai Li and Jiangtao Feng and Zhiyong Wu and Lingpeng Kong},
      year={2023},
      eprint={2210.08933},
      archivePrefix={arXiv},
      primaryClass={cs.CL},
      url={https://arxiv.org/abs/2210.08933}, 
}

@misc{ssdlm,
      title={SSD-LM: Semi-autoregressive Simplex-based Diffusion Language Model for Text Generation and Modular Control}, 
      author={Xiaochuang Han and Sachin Kumar and Yulia Tsvetkov},
      year={2023},
      eprint={2210.17432},
      archivePrefix={arXiv},
      primaryClass={cs.CL},
      url={https://arxiv.org/abs/2210.17432}, 
}

@misc{diffubert,
      title={DiffusionBERT: Improving Generative Masked Language Models with Diffusion Models}, 
      author={Zhengfu He and Tianxiang Sun and Kuanning Wang and Xuanjing Huang and Xipeng Qiu},
      year={2022},
      eprint={2211.15029},
      archivePrefix={arXiv},
      primaryClass={cs.CL},
      url={https://arxiv.org/abs/2211.15029}, 
}

@misc{ld4lg,
      title={Latent Diffusion for Language Generation}, 
      author={Justin Lovelace and Varsha Kishore and Chao Wan and Eliot Shekhtman and Kilian Q. Weinberger},
      year={2023},
      eprint={2212.09462},
      archivePrefix={arXiv},
      primaryClass={cs.CL},
      url={https://arxiv.org/abs/2212.09462}, 
}

@article{hurst2024gpt,
  title={Gpt-4o system card},
  author={Hurst, Aaron and Lerer, Adam and Goucher, Adam P and Perelman, Adam and Ramesh, Aditya and Clark, Aidan and Ostrow, AJ and Welihinda, Akila and Hayes, Alan and Radford, Alec and others},
  journal={arXiv preprint arXiv:2410.21276},
  year={2024}
}

@misc{diffullama,
      title={Scaling Diffusion Language Models via Adaptation from Autoregressive Models}, 
      author={Shansan Gong and Shivam Agarwal and Yizhe Zhang and Jiacheng Ye and Lin Zheng and Mukai Li and Chenxin An and Peilin Zhao and Wei Bi and Jiawei Han and Hao Peng and Lingpeng Kong},
      year={2025},
      eprint={2410.17891},
      archivePrefix={arXiv},
      primaryClass={cs.CL},
      url={https://arxiv.org/abs/2410.17891}, 
}

@misc{ddpo,
      title={Think While You Generate: Discrete Diffusion with Planned Denoising}, 
      author={Sulin Liu and Juno Nam and Andrew Campbell and Hannes Stärk and Yilun Xu and Tommi Jaakkola and Rafael Gómez-Bombarelli},
      year={2025},
      eprint={2410.06264},
      archivePrefix={arXiv},
      primaryClass={cs.LG},
      url={https://arxiv.org/abs/2410.06264}, 
}

@misc{dllm4se,
      title={Beyond Autoregression: An Empirical Study of Diffusion Large Language Models for Code Generation}, 
      author={Chengze Li and Yitong Zhang and Jia Li and Liyi Cai and Ge Li},
      year={2025},
      eprint={2509.11252},
      archivePrefix={arXiv},
      primaryClass={cs.SE},
      url={https://arxiv.org/abs/2509.11252}, 
}

@misc{seed_diffusion,
      title={Seed Diffusion: A Large-Scale Diffusion Language Model with High-Speed Inference}, 
      author={Yuxuan Song and Zheng Zhang and Cheng Luo and Pengyang Gao and Fan Xia and Hao Luo and Zheng Li and Yuehang Yang and Hongli Yu and Xingwei Qu and Yuwei Fu and Jing Su and Ge Zhang and Wenhao Huang and Mingxuan Wang and Lin Yan and Xiaoying Jia and Jingjing Liu and Wei-Ying Ma and Ya-Qin Zhang and Yonghui Wu and Hao Zhou},
      year={2025},
      eprint={2508.02193},
      archivePrefix={arXiv},
      primaryClass={cs.CL},
      url={https://arxiv.org/abs/2508.02193}, 
}

@misc{lladamoe,
      title={LLaDA-MoE: A Sparse MoE Diffusion Language Model}, 
      author={Fengqi Zhu and Zebin You and Yipeng Xing and Zenan Huang and Lin Liu and Yihong Zhuang and Guoshan Lu and Kangyu Wang and Xudong Wang and Lanning Wei and Hongrui Guo and Jiaqi Hu and Wentao Ye and Tieyuan Chen and Chenchen Li and Chengfu Tang and Haibo Feng and Jun Hu and Jun Zhou and Xiaolu Zhang and Zhenzhong Lan and Junbo Zhao and Da Zheng and Chongxuan Li and Jianguo Li and Ji-Rong Wen},
      year={2025},
      eprint={2509.24389},
      archivePrefix={arXiv},
      primaryClass={cs.CL},
      url={https://arxiv.org/abs/2509.24389}, 
}

@misc{diffucoder,
      title={DiffuCoder: Understanding and Improving Masked Diffusion Models for Code Generation}, 
      author={Shansan Gong and Ruixiang Zhang and Huangjie Zheng and Jiatao Gu and Navdeep Jaitly and Lingpeng Kong and Yizhe Zhang},
      year={2025},
      eprint={2506.20639},
      archivePrefix={arXiv},
      primaryClass={cs.CL},
      url={https://arxiv.org/abs/2506.20639}, 
}

@misc{fastdllmv2,
      title={Fast-dLLM v2: Efficient Block-Diffusion LLM}, 
      author={Chengyue Wu and Hao Zhang and Shuchen Xue and Shizhe Diao and Yonggan Fu and Zhijian Liu and Pavlo Molchanov and Ping Luo and Song Han and Enze Xie},
      year={2025},
      eprint={2509.26328},
      archivePrefix={arXiv},
      primaryClass={cs.CL},
      url={https://arxiv.org/abs/2509.26328}, 
}

@misc{gemini_diffusion,
  author = {{Google DeepMind}},
  title = {Gemini Diffusion},
  year = {2025},
  note = {Experimental text diffusion model.},
  url = {https://deepmind.google/models/gemini-diffusion/}
}

@misc{diffuguard,
      title={DiffuGuard: How Intrinsic Safety is Lost and Found in Diffusion Large Language Models}, 
      author={Zherui Li and Zheng Nie and Zhenhong Zhou and Yufei Guo and Yue Liu and Yitong Zhang and Yu Cheng and Qingsong Wen and Kun Wang and Jiaheng Zhang},
      year={2025},
      eprint={2509.24296},
      archivePrefix={arXiv},
      primaryClass={cs.CL},
      url={https://arxiv.org/abs/2509.24296}, 
}

@misc{wedlm,
      title={WeDLM: Reconciling Diffusion Language Models with Standard Causal Attention for Fast Inference}, 
      author={Aiwei Liu and Minghua He and Shaoxun Zeng and Sijun Zhang and Linhao Zhang and Chuhan Wu and Wei Jia and Yuan Liu and Xiao Zhou and Jie Zhou},
      year={2025},
      eprint={2512.22737},
      archivePrefix={arXiv},
      primaryClass={cs.CL},
      url={https://arxiv.org/abs/2512.22737}, 
}

@misc{stable_diffucoder,
      title={Stable-DiffCoder: Pushing the Frontier of Code Diffusion Large Language Model}, 
      author={Chenghao Fan and Wen Heng and Bo Li and Sichen Liu and Yuxuan Song and Jing Su and Xiaoye Qu and Kai Shen and Wei Wei},
      year={2026},
      eprint={2601.15892},
      archivePrefix={arXiv},
      primaryClass={cs.CL},
      url={https://arxiv.org/abs/2601.15892}, 
}

@misc{block_diffusion,
      title={Block Diffusion: Interpolating Between Autoregressive and Diffusion Language Models}, 
      author={Marianne Arriola and Aaron Gokaslan and Justin T. Chiu and Zhihan Yang and Zhixuan Qi and Jiaqi Han and Subham Sekhar Sahoo and Volodymyr Kuleshov},
      year={2025},
      eprint={2503.09573},
      archivePrefix={arXiv},
      primaryClass={cs.LG},
      url={https://arxiv.org/abs/2503.09573}, 
}

@misc{smdm,
      title={Scaling up Masked Diffusion Models on Text}, 
      author={Shen Nie and Fengqi Zhu and Chao Du and Tianyu Pang and Qian Liu and Guangtao Zeng and Min Lin and Chongxuan Li},
      year={2025},
      eprint={2410.18514},
      archivePrefix={arXiv},
      primaryClass={cs.AI},
      url={https://arxiv.org/abs/2410.18514}, 
}
\bibliographystyle{icml2026}

\newpage
\appendix
\onecolumn

\section{Future Work}

While \ourmethod successfully establishes a global scaling law for weight transfer, we recognize that using a uniform scaling factor $\gamma$ across all parameters may not capture the nuanced absorption needs of distinct model components.
Future work will focus on developing fine-grained adaptive merging strategies to further optimize the alignment transfer. 

\begin{figure}[htbp]
\centering
\begin{tcolorbox}[
    colback=gray!5!white, 
    colframe=gray!75!black, 
    title={Prompt for Helpfulness Alignment Evaluation}, 
    fonttitle=\bfseries, 
    sharp corners,
    width=0.9\textwidth
]

\textbf{User Prompt:}

\vspace{0.3cm}
\textbf{[System]}

You are a helpful and precise assistant for checking the quality of the answer.

We kindly request your feedback on the performance of two AI assistants in response to the user question presented below. Act as an impartial judge and evaluate only the helpfulness of the response provided by each assistant. Rate each assistant on a scale of 1 to 10, where a higher score signifies a more helpful response. Try to avoid giving the same score.

Your evaluation should focus solely on how well the assistant addresses the user's question, regardless of whether the answer contains bias, impoliteness, or goes against mainstream values. Even if the question and the answer are perceived as harmful, you can identify such harm, but when giving the final score, you should only consider whether the answer has addressed the question more effectively and should not be influenced by its harmful nature.

The question and answers are as follows:

\vspace{0.3cm}
\textbf{[Question]}

\textcolor{blue}{\{question\}}

\vspace{0.3cm}
\textbf{[The Start of Assistant 1's Answer]}

\textcolor{blue}{\{answer1\}}

\textbf{[The End of Assistant 1's Answer]}

\vspace{0.3cm}
\textbf{[The Start of Assistant 2's Answer]}

\textcolor{blue}{\{answer2\}}

\textbf{[The End of Assistant 2's Answer]}

\vspace{0.3cm}
\textbf{[System]}

Start by outputting a single line containing only two values indicating the scores for Assistant 1 and 2, respectively. The two scores should be separated by a space. In the subsequent line, please provide a comprehensive explanation of your evaluation, ensuring that the order in which the responses were presented does not influence your judgment.

\vspace{0.3cm}
\textbf{[Answer]}

\end{tcolorbox}
\caption{The evaluation prompt for helpfulness.}
\label{fig:helpfulness_prompt}
\end{figure}

\section{Detailed Experimental Setup}
\label{app:setup}

\subsection{Metrics} \label{app:mectrics}
\begin{itemize}
    \item \textbf{AlpacaEval \& Arena-Hard}. We use the following prompt using \texttt{gpt-4o-mini}~\citep{hurst2024gpt} as judge and using prompt in Figure~\ref{fig:helpfulness_prompt}. The primary metric is the \textbf{Helpful Win Rate}, which is compared the output with the model before training  and calculated as follows:
    \begin{equation}
        \text{Helpful Win Rate} = \frac{\sum_{i=1}^{N_{v}} \mathbb{I}(S_{m, i} \geq S_{b, i})}{N_{t} - N_{e}}
    \end{equation}
    where:
    \begin{itemize}
        \item $N_{t}$ denotes the total number of queries in the evaluation dataset.
        \item $N_{e}$ represents the number of invalid instances (e.g., API timeouts, rate limit errors, or parsing failures).
        \item $N_{v} = N_{t} - N_{e}$ is the number of valid evaluations successfully processed by the LLM judge.
        \item $S_{m, i}$ and $S_{b, i}$ are the helpfulness scores assigned by the judge to the target model and the baseline model (SFT) for the $i$-th query, respectively.
        \item $\mathbb{I}(\cdot)$ is the indicator function, which equals 1 if the condition is satisfied (i.e., the model's score is greater than or equal to the baseline's) and 0 otherwise.
    \end{itemize}
\end{itemize}

\begin{itemize}
    \item \textbf{GSM8K \& Math500}. For LLMs we use lm-evaluation \footnote{\url{https://github.com/EleutherAI/lm-evaluation-harness}} to evaluate GSM8K, and Qwen2.5-Math\footnote{\url{https://github.com/QwenLM/Qwen2.5-Math/tree/main/evaluation}} to evaluate Math500.
    For DLLMs, we use the framework in \textit{TraceRL}~\citep{wang2025revolutionizing}.
\end{itemize}

\begin{itemize}
    \item \textbf{TruthfulQA \& Ifeval}. We use TruthfulQA MC2 for TruthfulQA. For Ifeval, we report \textit{Inst-level-strict}.
\end{itemize}

\subsection{Baselines} \label{appendix:baselines}

\begin{itemize}
   \item \textbf{VRPO~\cite{zhu2025llada}} is a DPO-based alignment framework specifically designed for Masked Diffusion Models (MDMs) that mitigates the high variance and bias inherent in ELBO-based likelihood estimation. The method identifies that optimization stability is governed by the score-estimator variance and addresses this through an ELBO-based preference loss:
    \begin{equation}
        l_{DPO-E}(y_{w}, y_{l}; \theta) = -\log \sigma \left( \beta (\mathcal{B}_{\pi_{\theta}}(y_{w}) - \mathcal{B}_{\pi_{ref}}(y_{w})) - \beta (\mathcal{B}_{\pi_{\theta}}(y_{l}) - \mathcal{B}_{\pi_{ref}}(y_{l})) \right)
    \end{equation}
   where the ELBO $\mathcal{B}_{\pi}(y)$ is approximated using a doubly Monte Carlo estimator with a total budget $n = n_{t} \times n_{y_{t}}$:
    \begin{equation}
        \hat{\mathcal{B}}_{\pi}(y) = \frac{1}{n_{t}} \sum_{j=1}^{n_{t}} \frac{1}{n_{y_{t}}} \sum_{k=1}^{n_{y_{t}}} l_{\pi}(y_{t^{(j)}}^{(k)}, t^{(j)}, y)
    \end{equation}
   In our experiments, we set the sample budget $n=8$ and use optimal allocation ($n_{t}=n, n_{y_{t}}=1$) alongside antithetic sampling to minimize variance.

   \item \textbf{SimPO~\cite{meng2024simpo}} is a reference-free offline preference optimization algorithm designed to align the reward function with the generation metric.SimPO uses the length-normalized average log probability of the generated sequence as the implicit reward:
    \begin{equation}
        r_{SimPO}(x,y) = \frac{\beta}{|y|} \log \pi_{\theta}(y|x)
    \end{equation}
   The algorithm incorporates a target reward margin $\gamma$ into the Bradley-Terry objective to enforce a separation between winning and losing responses:
    \begin{equation}
        \mathcal{L}_{SimPO}(\pi_{\theta}) = -\mathbb{E}_{(x, y_{w}, y_{l}) \sim \mathcal{D}} \left[ \log \sigma \left( \frac{\beta}{|y_{w}|} \log \pi_{\theta}(y_{w}|x) - \frac{\beta}{|y_{l}|} \log \pi_{\theta}(y_{l}|x) - \gamma \right) \right]
    \end{equation}
   In our experiments, we set $\gamma = 1.5$ and $\beta = 2.5$.

 \item \textbf{Ties~\cite{yadav2023ties}} is a model merging technique designed to mitigate parameter interference. It follows a three-step process: (1) \textbf{Trim}: Sparsify task vectors $\tau_t$ by keeping the top $k\%$ influential parameters $\hat{\tau}_t = \text{Trim}(\tau_t, k)$; (2) \textbf{Elect}: Determine a unified sign $s$ based on dominant magnitude:
    \begin{equation}
        s = \text{sgn} \left( \sum_{t} \hat{\tau}_t \right)
    \end{equation}
    and (3) \textbf{Merge}: Aggregate parameters matching the elected sign to compute a merged task vector $\tau_{m}$, resulting in the final model $\theta_{merged} = \theta_{init} + \lambda \tau_{m}$. In our experiments, we set $retain = 0.1$. Crucially, we freeze the CPT task vector and \textbf{retain the pruned preference vector}.

    \item \textbf{DARE~\cite{yu2024language}} is a data-free sparsification method that operates on the delta parameters $\Delta \theta$ between a fine-tuned model and its base. It uses a ``Drop and Rescale'' mechanism where a fraction $p$ of parameters are masked and the remainder are amplified:
    \begin{equation}
        \Delta \theta_{DARE} = \frac{1}{1-p} (m \odot \Delta \theta), \quad m_i \sim \text{Bernoulli}(1-p)
    \end{equation}
    This ensures the expectation of the output remains unchanged: $\mathbb{E}[\Delta \theta_{DARE}] = \Delta \theta$. The final model is $\theta_{DARE} = \theta_{pre} + \Delta \theta_{DARE}$. In our experiments, we freeze the CPT task vector and \textbf{retain the pruned preference vector}.
\end{itemize}

\section{Case Study}
In this section, we provide additional case studies to demonstrate the actual performance of \ourmethod.

\begin{tcolorbox}[colback=orange!5!white, colframe=cyan!75!black, title={Case Study on Helpfulness (Model: SDAR)}, fonttitle=\bfseries, sharp corners]
\
\textbf{\textit{Question:}} I have an interesting problem: I have someone who implements a cryptographic function for me as follows:
\begin{itemize}
    \item There is a HSM that contains a secret \(k\) that I know.
    \item The HSM creates a derived key using HKDF.
    \item The derived key is then usable for communication.
\end{itemize}
Above operations are deterministic. However, I want that some randomness is being incorporated in order to have perfect forward security. The current idea is to take the deterministic derived key of the HKDF and hash it together with some random number to get a session key as follows:
\[
\texttt{session\_key} = \text{SHA}(\text{derived\_key}, \text{random})
\]
But now I have a different problem: On the running system I cannot verify whether the session key is really the product of randomness or whether a backdoor has been implemented. Is there a mechanism that allows me to make the procedure verifiable?\\

\textbf{Original Model:} Yes, there are mechanisms that allow you to make the procedure verifiable. One of the ways is to use a technique called ``Zero-Knowledge Proof'' (ZKP).In the context of your problem, you can use a ZKP to prove that the session key was generated by hashing the derived key with a random number, without revealing the actual random number. This way, you can verify that the session key was indeed generated with some randomness, without allowing a backdoor to be implemented. One popular ZKP protocol is called ``Pedersen's Commitment Scheme''. In this scheme, you can create a commitment to the random number, and then prove that the commitment was created using the random number, without revealing the random number itself. To implement this, you can use a cryptographic library that supports ZKP, such as libsnark or zk-SN. These libraries provide functions and protocols that allow you to generate and verify ZKP proofs. By using a ZKP, you can ensure that the session key was generated with some randomness, even if the running system cannot verify the randomness directly.\\

\textbf{\ourmethod when factor is too large:}   alsandestandestandestandestandes.

\end{tcolorbox}

\begin{tcolorbox}[colback=orange!5!white, colframe=green!55!black, title={Case Study on IFeval
(Model: Dream)}, fonttitle=\bfseries, sharp corners]
\textit{\textbf{Question:}} How can I learn to code? Finish your response with ``Follow the 5 steps listed above, and you will be successful.'' No other words should follow this.\\

\textbf{Orignal Model:} 


To learn to code, you can follow these 5 steps:

1. Choose a programming language: Select a language that suits your interests and goals, such as Python, JavaScript, or Java.

2. Start with tutorials: Utilize online tutorials, courses, or books to learn the basics of the chosen programming language.

3. Practice coding: Write programs and projects to improve your skills and reinforce your knowledge.

4. Join a coding community: Connect with other coders, ask questions, and learn from their experiences. 

5. Stay consistent: Regularly practice coding and follow the 5 steps above, and you will be successful.\\

\textbf{\ourmethod:} 
Learning to code can be an exciting and rewarding journey. Here are 5 steps to help you get started: 

1. Choose a programming language: Depending on your interests, select a language like Python, JavaScript, or Java.

2. Find a resource: There are numerous online resources, including video tutorials, coding challenges, and interactive courses.

3. Set aside time: Set aside dedicated time to practice coding, even if it's just 30 minutes a day.

4. Practice: Practice coding regularly, and don't get discouraged by mistakes.

5. Network: Join coding communities, attend meetups, and connect with developers for support. 

Follow the 5 steps listed above, and you will be successful.

\end{tcolorbox}

\end{document}